\documentclass{article}
\usepackage[final]{neurips_2025}
\usepackage{newtxtext} 

\usepackage[utf8]{inputenc} 
\usepackage[T1]{fontenc}    
\usepackage{comment}
\usepackage{amsthm}

\usepackage{hyperref}       
\usepackage{url}            
\usepackage{booktabs}       
\usepackage{amsfonts}       
\usepackage{nicefrac}       
\usepackage{microtype}      

\title{LSAM: Asynchronous Distributed Training with Landscape-Smoothed Sharpness-Aware Minimization}

\author{%
  Yunfei Teng \\
  yfteng@pku.edu.cn \\
  \And
  Sixin Zhang \\
  sixin.zhang@irit.fr \\
}

\usepackage{amsmath, amssymb, amsthm, mathtools, bm}

\newcommand{\E}{\mathbb{E}}

\newcommand{\R}{\mathbb{R}}
\newcommand{\norm}[1]{\lVert #1 \rVert}
\newcommand{\eps}{\varepsilon}

\usepackage{wrapfig}
\usepackage{mdframed}
\usepackage{enumitem}

\usepackage{svg}
\usepackage{xcolor}
\usepackage{colortbl}
\usepackage{graphicx}
\usepackage{subcaption}
\usepackage{multirow}
\usepackage{makecell}
\usepackage{threeparttable}
\definecolor{forestgreen}{RGB}{34,139,34}

\usepackage{float}
\usepackage{algorithm}
\usepackage{algpseudocode}

\usepackage{newfloat}
\usepackage{listings}
\DeclareCaptionStyle{ruled}{labelfont=normalfont,labelsep=colon,strut=off}
\floatstyle{ruled}
\newfloat{listing}{tb}{lst}{}
\floatname{listing}{Listing}

\usepackage{natbib}
\setcitestyle{authoryear}

\usepackage{cleveref}
\usepackage{appendix}

\newtheorem{theorem}{Theorem}
\newtheorem{lemma}[theorem]{Lemma}
\newtheorem{corollary}[theorem]{Corollary}
\newtheorem{proposition}[theorem]{Proposition}

\theoremstyle{definition}
  
  \newtheorem{assumption}{Assumption}
\theoremstyle{remark}
  \newtheorem{remark}{Remark}

\crefname{theorem}{theorem}{theorems}
\crefname{lemma}{lemma}{lemmas}
\crefname{proposition}{proposition}{propositions}
\crefname{corollary}{corollary}{corollaries}
\crefname{assumption}{assumption}{assumptions}
\crefname{definition}{definition}{definitions}
\crefname{remark}{remark}{remarks}

\newcommand{\sixin}[1]{{\color{black}#1}}

\begin{document}
\maketitle
\vspace{-0.2in}

\begin{abstract}
\vspace{-0.1in}
While Sharpness-Aware Minimization (SAM) improves generalization in deep neural networks by minimizing both loss and sharpness, it suffers from inefficiency in distributed large-batch training. We present \emph{Landscape-Smoothed} SAM (LSAM), a novel optimizer that preserves SAM’s generalization advantages while offering superior efficiency. LSAM integrates SAM’s adversarial steps with an \emph{asynchronous} distributed sampling strategy, generating an asynchronous distributed sampling scheme, producing a smoothed sharpness-aware loss landscape for optimization. This design eliminates synchronization bottlenecks, accelerates large-batch convergence, and delivers higher final accuracy compared to data-parallel SAM.
\end{abstract}
\vspace{-0.1in}

\vspace{-0.1in}
\section{Introduction}
\vspace{-0.1in}
Recent advancements in computer vision and natural language processing have been significantly fueled by meticulously designed optimization methods. In this study, we address the optimization challenge expressed as:
$
f(x) = \min_{x \in \mathbb{R}^d} \, \mathbb{E}_{\xi \sim \mathcal{D}}\left[ f(x;\xi) \right] \approx \frac{1}{n} \sum_{i=1}^n f(x;\xi_i),
$
where $f(\cdot;\xi) \colon \mathbb{R}^d \to \mathbb{R}$ is a differentiable (though potentially nonconvex) loss function. Here, $\xi$ is sampled from an unknown distribution $\mathcal{D}$ (typically representing the data distribution), and $x$ denotes the trainable model parameters. In practice, the objective is often approximated by minimizing the empirical risk over a finite dataset instead. This formulation captures a wide range of machine learning tasks, including least-squares regression, neural network training and simulation of molecular dynamics. 

It is widely recognized that deep neural networks are highly nonconvex and challenging to optimize. \emph{Sharpness-Aware Minimization} (SAM) \citep{foret2020sharpnessaware} improves generalization by simultaneously minimizing the empirical loss and the local \emph{sharpness} of the loss landscape. For model parameters $x$ and objective function $f$, it first finds the adversarial perturbation ${\epsilon}(x)=\arg\max_{\|{\epsilon}\|\le\rho} f(x+{\epsilon})$, then applies a conventional optimizer to minimise the worst-case loss $f(x+{\epsilon}(x))$, guiding the parameters into wider, flatter minima and thus enhancing generalization.

While SAM delivers significant generalization gains, these improvements deteriorate rapidly as the batch size grows \citep{pmlr-v162-andriushchenko22a,wen2023how,FSAM}. In practice, one therefore resorts to a mini-batch surrogate as $\mathbb{E}_{\xi \sim \mathcal{D}} \left[f\left( x + \epsilon(x); \xi \right) \right]$,
Consequently, confining SAM's effectiveness to moderate batch sizes (e.g., 128) rather than large-batch regimes. This batch sensitivity fundamentally limits SAM's scalability in conventional data-parallel distributed settings. Under data parallelism, adding workers forces an unfavorable trade-off: either shrink per-worker batches or balloon global batch sizes—both undermining SAM's training dynamics and restricting its utility for large-scale distributed training.

\begin{wrapfigure}{l}{0.5\textwidth} 
\vspace{-0.24in}
\begin{mdframed}[linewidth=0.6pt, roundcorner=4pt]
Thus, beyond the conventional \emph{data-parallel} scheme, is there any other strategy can enable large-scale distributed training of SAM?
\end{mdframed}
\vspace{-0.25in}
\end{wrapfigure}

This paper answers that question by introducing \emph{sampling parallelism}—a paradigm that overcomes SAM’s batch‐size sensitivity and scales gracefully in distributed settings. Sampling parallelism thus establishes a new conceptual framework that complements existing data parallelism strategies. Although earlier methods such as \citep{easgd,entropysgd,PARLE,lsgd,GRAWA} can be retroactively viewed through this lens, they did not explicitly formulate the idea, nor did they provide optimally tuned implementations.

To establish connections between SAM and modern sampling techniques, we first introduce the \emph{Boltzmann distribution}~\citep{gibbs1902elementary}. This fundamental distribution assigns state probabilities that decay exponentially with energy. In machine learning contexts, where energy corresponds to a loss function $f$, this yields:
\begin{equation}
\pi_{\text{SAM}} (x) \propto \exp\left(-\mathbb{E}_{\xi \sim \mathcal{D}} \left[f\left( x + \epsilon(x); \xi \right) \right]\right).
\label{eq:sam_full}
\end{equation}
This formulation suggests that low-loss states occur with higher probabilities. Later in this paper, we will demonstrate that it constitutes a valid distribution under mild constraints.

Prior work \citep{pmlr-v178-chen22c,pmlr-v134-lee21a,entropysgd,huang2024reverse} demonstrates that convolving the loss landscape with a Gaussian kernel, coupled with appropriate regularization, offers an effective optimization solution. Building on this foundation, we introduce \emph{Landscape-Smoothed} SAM (LSAM)—a novel optimizer specifically designed for scalable and robust distributed learning. LSAM simultaneously enhances the smoothness of landscape and strengthens optimization stability through entropy-enhanced regularization. Our theoretical and empirical analyses confirm LSAM's compatibility with asynchronous operation, enabling a highly efficient training paradigm.

We contribute: (1) A novel \emph{sampling-parallel} framework for distributed SAM that prevents batch inflation; (2) Demonstration that LSAM unifies SAM's flat minima and Gaussian smoothing's deep minima; (3) Theoretical proof that LSAM matches SGD convergence rates; (4) Empirical validation showing LSAM's superiority over data-parallel SAM and similar distributed methods.

\vspace{-0.1in}
\section{Related Work}
\vspace{-0.1in}
Classical sampling approaches such as Langevin dynamics~\citep{max2011}, MALA~\citep{grenanderRepresentationsKnowledgeComplex1994}, and HMC~\citep{nealHMC2011} couple gradients with injected noise, while proximal and entropy-smoothed variants enhance scalability in high dimensions~\citep{pmlr-v178-chen22c,pmlr-v134-lee21a,chen2024gibbs}. Flatness-oriented optimization evolved from the early observation that wide minima improve generalization~\citep{hochreiter1997flat,keskar2017large} to smoothing techniques such as Entropy-SGD~\citep{entropysgd}, diffusion-based perturbations~\citep{zhu2022diffusion,wen2023sharpnessaware}, and augmentation strategies~\citep{zhang2018mixup}, as well as adversarial formulations like SAM~\citep{foret2020sharpnessaware} and its extensions~\citep{kwon2021asam,kim2023smooth,qu2023flatsam}. Theory has further tied flatness to PAC-Bayes bounds and spectral properties of the loss~\citep{neyshabur2017exploring,andriushchenko2023tensor}. Efforts to scale SAM efficiently include stochastic perturbations, periodic updates, and adaptive learning rates~\citep{efsam,looksam,adasam,bmsam}, with AsyncSAM~\citep{asyncsam} overlapping perturbation and update steps, and federated variants adapting SAM to heterogeneous data~\citep{fedsam,fedsmoo,fedlesam,fedgloss}.

\begin{remark}
Our approach differs by unifying sampling with SAM through communication-level asynchrony, yielding the first sampling-driven framework for distributed sharpness-aware optimization. Nevertheless, our method still requires double the communication cost, same as the original SAM.
\end{remark}

\vspace{-0.1in}
\section{Theoretical Analysis}
\label{sec:theory}

\vspace{-0.05in}
\subsection{Formulation}
\label{sec:form}
\vspace{-0.05in}
In line with the work of \citep{foret2020sharpnessaware}, we formulate the sharpness-aware minimization (SAM) objective as follows:
\begin{equation}
f_{\text{SAM}}^{\text{max}}(x) = \max_{\|\epsilon\| \leq \rho} f(x + \epsilon) 
\end{equation}
This objective seeks to minimize not just the loss at the current parameters $x$, but the maximum loss within a small $\ell_2$-ball of radius $\rho$ around $x$, thereby promoting solutions in flatter regions of the loss landscape that are more robust to perturbations and tend to generalize better.

In practical implementations, the maximization over $\epsilon$ is often approximated using a first-order Taylor expansion, leading to the perturbation
$\epsilon = \rho \cdot \frac{\nabla f(x)}{\|\nabla f(x)\|}$ 
(for the $\ell_2$-norm case).
To enhance numerical stability and avoid issues like division by zero when the gradient norm is small, this is adjusted to $f_{\text{SAM}}^{\text{max}}$ is modified to:
\vspace{-0.1in}
\begin{equation}
f_{\text{SAM}}(x) = f\left(x+\frac{\rho\,\nabla f(x)}{\lVert\nabla f(x)\rVert+\gamma}\right),
\end{equation}
where $\gamma > 0$ is a small regularization constant. This approximation allows efficient computation via a single additional forward-backward pass to evaluate the perturbed loss, which is then used to compute the gradient for the parameter update.

Building on this, the approximation naturally induces a Boltzmann distribution over the parameters $x$:
\begin{equation}
\pi_{\text{SAM}}(x) \propto \exp\left(-f_{\text{SAM}}(x)\right),
\end{equation}
assigning higher probability to parameter vectors with lower sharpness-aware loss. Under mild assumptions defined and normalizable, linking SAM to sampling-based techniques in optimization and generative modeling.

\begin{theorem}[Score Estimation]\label{lma:score}
Let $f:\R^{d}\to\R$ and $k:\R^{d}\times\R^{d}\to [0,\infty)$ satisfy  
\Cref{thm:ass_fx,ass:kernel_slice} (in \Cref{app:additional_theorems}), and define
\[
I(y) := \int_{\R^{d}} e^{-f(T_{\rho,\gamma}(x)) - k(x,y)} \, dx, 
\quad \pi_{\text{LSAM}}(y) := \tfrac{I(y)}{Z_{\rho,\gamma} Z}.
\]
Assume:  
(i) $\nabla_y k(x,y)$ exists $\forall (x,y)\in\R^d\times\R^d$;  
(ii) $f$ is bounded below;  
(iii) $\|\nabla_y k(x,y)\| \le g(x)$ for some integrable $g:\R^d\to\R$.  
Then differentiation under the integral is valid, and
\begin{align}
\nabla_{y}\log \pi_{\text{LSAM}}(y) = -\,\E_{x\sim q(\cdot|y)}[\nabla_{y}k(x,y)], \quad
q(x|y) = \tfrac{e^{-f(T_{\rho,\gamma}(x)) - k(x,y)}}{I(y)}. \label{eq:cond_esam}
\end{align}
\noindent Thus, the log-density gradient equals the negative posterior mean of $\nabla_y k(x,y)$ under $q$.
\end{theorem}

\begin{remark}
Our contribution generalizes prior work~\citep{huang2024reverse,chen2024gibbs,entropysgd} by broadening their findings to encompass general kernels, extending beyond the Gaussian setting. The key observation is that the score at any point $y$ arises from kernel-weighted averages over contributions from its neighbor.
\end{remark}

Following prior work~\citep{entropysgd}, we adopt a \emph{Gaussian kernel} in our implementation. Under this formulation, $\pi_{\text{LSAM}}$ integrates the adversarial robustness of SAM—which encourages flatter minima by maximizing over worst-case perturbations—with kernel smoothing. The latter alleviates irregularities in the optimization landscape by convolving the loss surface.

\begin{wrapfigure}{r}{0.5\textwidth} 
\begin{minipage}{\linewidth}
\vspace{-0.5in}
\begin{algorithm}[H]
\caption{One-step LSAM / ESGD Update}
\label{alg:dual_g}
Select a coupling parameter $\lambda \geq 0$, a mixing coefficient $\alpha \in (0, 1]$, and a sequence of step sizes $\eta_t = \eta_0 / \sqrt{t+1}$ for some $\eta_0 > 0$. The algorithm produces the following updates:
\begin{align}
x_{t+1} &= x_t - \eta_t (g_t + \lambda (x_t - y_t)), \label{eq:esam1}\\
y_{t+1} &= \alpha x_{t+1} + (1 - \alpha) y_t, \label{eq:esam2}
\end{align}
where $g_t$ is a stochastic gradient which depends on an i.i.d random sample $\xi_t \sim \mathcal{D}$.
For convenience, we define:
\vspace{-0.05in}
$$\begin{aligned}
G_t &:= \nabla f(x_t) + \lambda (x_t - y_t).
\end{aligned}$$
\textit{Comment.} Informally, repeatedly applying Equations~\Cref{eq:esam1,eq:esam2} in alternation with appropriate injected noise produces updates analogous to a discrete-time Gibbs sampler. In contrast, for optimization we perform each update just once in a sequential recursive schedule.
\end{algorithm}
\vspace{-0.55in}
\end{minipage}
\end{wrapfigure}

\begin{remark}
Despite its sharpness-aware objective, SAM is often used as merely a first-order optimizer like gradient descent, potentially susceptible to flat yet shallow minima. We demonstrate, however, that SAM fundamentally reshapes the probability landscape. When integrated into our framework as LSAM—which implicitly incorporates zero-order information—SAM can be strategically guided to deeper minima, transcending its traditional role as a standalone optimizer.
\end{remark}

\vspace{-0.15in}
\subsection{Convergence Rate}
\label{sec:opt}
\vspace{-0.1in}
In this section, we analyze the convergence behavior of LSAM towards the minima. We first present the algorithmic framework, then derive convergence guarantees for \text{ESGD}, and ultimately show that \text{LSAM} attains the same convergence rate as classical SGD in approaching stationary points of the objective $f$.

We cast the optimization procedure an alternating update scheme driven by stochastic gradients. The iterative routines for both ESGD and LSAM are characterized in \Cref{alg:dual_g}. Then, under \Cref{asm:oracle} and through \Cref{alg:dual_g}, we prove $G_t$ convergence for both: (i) ESGD without perturbation (\Cref{thm:convergence_esgd}), and (ii) LSAM with constant and decaying perturbations (\Cref{thm:convergence_constant_perturbation} and \Cref{thm:convergence_decaying_perturbation}, respectively). All proofs of this \Cref{sec:form,sec:opt} are deferred to \Cref{app:proofs}.

\setcounter{assumption}{2}
\begin{assumption}[Stochastic-gradient oracle]\label{asm:oracle}
Consider the following conditions:
\begin{enumerate}[label=\textup{(C\arabic*)},
                 itemsep=3pt,leftmargin=2.5em]
\item \textbf{Bounded variance.}  
      There exists \(\sigma\ge0\) such that
      $
        \mathbb{E}_\xi\bigl[\,
          \|f (x; \xi)-\nabla f(x)\|^{2}
        \bigr]
        \;\le\;\sigma^{2},\ \forall x \in \R^d.
      $
\item \textbf{\(L\)-smoothness.}  
      There exists \(L\ge0\) such that
      $
        \|\nabla f(x_1)-\nabla f(x_2)\|
        \;\le\;
        L\,\|x_1-x_2\|,\ \forall x_1,x_2.
      $
\item \textbf{Stochastic \(L\)-smoothness.} There exists $L \geq 0$ 
such that for a.e. $\xi$, $f(\cdot; \xi)$
satisfies $L$-smoothness in (C2)
$
  \|\nabla f(x_1;\xi)-\nabla f(x_2;\xi)\| \le L\|x_1-x_2\|, \; \forall x_1,x_2 \in \R^d.
$
 
\item \textbf{Bounded stochastic gradient norm.} 
The expected norm of the stochastic gradient is uniformly bounded
$
    \exists C \geq 0 \text{ such that } \; \mathbb{E}_{\xi} \|\nabla f(x; \xi)\| \leq C,\ \forall x \in \R^d.
$
\end{enumerate}
\end{assumption}

\begin{theorem}[Convergence of $G_t$ without Perturbation]
\label{thm:convergence_esgd}
Under (C1)-(C2) of \Cref{asm:oracle}, consider the updates with $g_t = \nabla f (x_t; \xi_t)$ with $\eta_0 \leq 1/ ( L + \lambda)$. Assume $f$ is bounded below. Then the average squared norm of $G_t$ satisfies
$
\frac{1}{T} \sum_{t=0}^{T-1} \mathbb{E}[\|G_t\|^2] = \mathcal{O}\left(\frac{\log T}{\sqrt{T}}\right).
$
\end{theorem}

\begin{theorem}[Convergence of $G_t$ with Constant Perturbation]
\label{thm:convergence_constant_perturbation}
Under (C1) - (C3) of \Cref{asm:oracle}, consider the updates with $g_t = \nabla f\left(x_t + \rho \frac{\nabla f(x_t; \xi_t)}{\|\nabla f(x_t; \xi_t)\| + \gamma}; \xi_t\right)$ with $\eta_0 \leq \frac{1}{4(L + \lambda)}$. Assume $f$ is bounded below and pick up a constant $\rho > 0$. Then the average squared norm of $G_t$ satisfies:
$
\frac{1}{T} \sum_{t=0}^{T-1} \mathbb{E}[\|G_t\|^2] \leq 4 L^2\rho^2 + \mathcal{O}\left(\frac{\log T}{\sqrt{T}}\right),
$
and thus converges to a neighborhood of size $\mathcal{O}(\rho^2)$.
\end{theorem}

\begin{theorem}[Convergence with Decaying Perturbation]
\label{thm:convergence_decaying_perturbation}
Under (C1)-(C3) of \Cref{asm:oracle}, consider the updates with $g_t = \nabla f\left(x_t + \rho_t \frac{\nabla f(x_t; \xi_t)}{\|\nabla f(x_t; \xi_t)\| + \gamma}; \xi_t\right)$ with $\eta_0 \leq \frac{1}{4(L + \lambda)}$. Assume $f$ is bounded below and let $\rho_t = \rho_0 / \sqrt{t+1}$ for some $\rho_0 > 0$. Then the average squared norm of $G_t$ satisfies:
$
\frac{1}{T} \sum_{t=0}^{T-1} \mathbb{E}[\|G_t\|^2] \leq \mathcal{O}\left(\frac{\log T}{\sqrt{T}}\right),
$
and thus converges to zero as $T \to \infty$.
\end{theorem}

\begin{wrapfigure}{r}{0.45\textwidth} 
\vspace{-0.45in}
\begin{minipage}{0.48\textwidth}
\begin{algorithm}[H]
\caption{Asynchronous Distributed LSAM}
\label{alg:esam}
\begin{flushleft}
\begin{algorithmic}[1]
\State \textbf{Inputs:} pulling coefficient $\lambda$, perturbation radius $\rho$, learning rates for sampling and optimization $(\eta, \eta')$, sync period $\tau$, and momentum factor $\beta$.
\State \textbf{Initialization:} Randomly initialize parameters for each worker $\{x_0^{(i)}\}_{i=1}^{n}$ and the global center $y_0$; set local counters $t_x^{(i)}\leftarrow 0$ and global counter $t_y\!\leftarrow 0$.
\While{not converged}
  \State \textcolor{gray}{\% \emph{Asynchronous sampling step}}
  \ForAll{$i=1,2,\dots,n$ \textbf{in parallel}}
    \State $s_{t_x^{(i)}}^{(i)} \!\gets\! \nabla_x\log q_{x|y}\!\left(x_{t_x^{(i)}}^{(i)};y_{t_y} \right)$
    \State $x_{t_x^{(i)}+1}^{(i)}
           \!\gets\!
           \textsc{Sampler}\!\left(x_{t_x^{(i)}}^{(i)},
           s_{t_x^{(i)}}^{(i)}, \eta\right)$
    \State $t_x^{(i)}\gets t_x^{(i)}+1$
  \EndFor
  \State
  \State \textcolor{gray}{\% \emph{Synchronous optimization step}}
  \If{$\displaystyle \sum_{i=1}^{n} t_x^{(i)}\bmod n\tau = 0$}
     \State $g'_{t_y}\gets
       \frac{1}{n\tau}\sum_{i=1}^{n}\sum_{s=1}^{t_x^{(i)}}x_{s}^{(i)} \;-\; y_{t_y}$
     \State $g_{t_y}\gets g'_{t_y} 
            \;+\; \beta\bigl(g'_{t_y}-g'_{t_y-1}\bigr)$
     \State $y_{t_y+1}\gets \textsc{Optimizer}\!\left(y_{t_y}, g_{t_y}, \eta' \right)$
     \State $t_y\gets t_y+1$;\quad reset all $t_x^{(i)}\gets 0$
  \EndIf
\EndWhile
\end{algorithmic}
\end{flushleft}
\end{algorithm}
\end{minipage}
\vspace{-0.8in}
\end{wrapfigure}

\Cref{thm:convergence_esgd,thm:convergence_constant_perturbation,thm:convergence_decaying_perturbation} guarantee the convergence of $\|G_t\|$ but not necessarily $\|\nabla f(x_t)\|$. Next, we require (C4) of \Cref{asm:oracle} and prove the corresponding result in \Cref{lem:anchor-gap}.

\begin{lemma}[Decay of the anchor gap]\label{lem:anchor-gap} 
Let (C1) - (C4) of \Cref{asm:oracle} hold, and let the iterates $\{(x_t,y_t)\}_{t\ge0}$ be generated by \Cref{alg:dual_g} with the $g_t$ in \Cref{thm:convergence_esgd} (resp. \Cref{thm:convergence_decaying_perturbation}). 
Set $\lambda > 0$ and choose $D \geq C/\lambda$
(resp. $D \geq  (C + \rho_0 L)/\lambda$).
Assume step sizes $\eta_t = \eta_0 / \sqrt{t+1}$ for $\eta_0 > 0$, and initial gap $\|z_0\| \leq D$, where $z_t := x_t - y_t$. Then, for all $t \geq 0$, the norm of $z_t$ remains bounded as $\|z_t\| \leq D$. Moreover,
$
\E[\|z_t\|^2] = \mathcal{O}(1/\sqrt{t}) \text{ and } \lim_{t \to \infty} \E[\|z_t\|^2] = 0.
$
\end{lemma}

\begin{corollary}[Gradient norm convergence]\label{cor:grad_norm_convergence}
Let (C1) - (C4) of \Cref{asm:oracle} hold and assume the conditions of \Cref{lem:anchor-gap} are satisfied. 
Then for \textsc{ESGD} (resp. \textsc{LSAM} with decaying perturbation) which satisfy the extra assumptions in \Cref{thm:convergence_esgd} (resp. \Cref{thm:convergence_decaying_perturbation}),
the expected squared gradient norm satisfies:
$
\frac{1}{T}\sum_{t=1}^T \mathbb{E}\left[\|\nabla f(x_t)\|^2\right] = \mathcal{O}\left(
\frac{\log T}{\sqrt{T}} \right).
$
\end{corollary}
\vspace{-0.05in}

SAM's convergence analysis shows distinct behaviors: it fails to converge to zero with a fixed $\rho$ stochastically, but converges with a decaying $\rho$. We prove that LSAM matches SAM's convergence rate for both constant adversarial parameters \citep{psam} and a decaying adversarial factor \citep{pmlr-v162-andriushchenko22a,FSAM}. 

\vspace{-0.2in}
\section{Algorithm}
\vspace{-0.1in}
In this section, we broaden the algorithmic structure presented in \Cref{alg:dual_g} to encompass a distributed optimization environment, by incorporating sampling and acceleration methodologies to proficiently locate the modes of $\pi_{\text{LSAM}}(y)$ that exhibit strong generalization capabilities. The dual-loop structure comprises two tightly coupled components: (i) An \emph{inner} sampling loop that leverages Langevin Markov chain Monte Carlo (MCMC) methods~\citep{max2011} to efficiently explore the parameter space and (ii) an \emph{outer} optimization loop that applies a gradient descent mechanism \citep{bottou-98x} to push the parameters toward optimal solutions. We elaborate on each component in the subsequent subsections.

\vspace{-0.05in}
\subsection{Langevin Dynamics for Sampling}
\vspace{-0.05in}
Since the score $\nabla \log \pi_{\text{LSAM}}(y)$ in \Cref{lma:score} does not admit a closed-form computation—owing to the intractable partition function and the convolution integral—we generate samples from the target density via MCMC techniques, chiefly Metropolis–Hastings~\citep{metropolis1953equation} or its stochastic-gradient counterpart, Stochastic Gradient Langevin Dynamics (SGLD)~\citep{max2011}. To alleviate the overhead associated with sampling, one could instead directly minimize a proximal approximation of the entropy-regularized objective using standard stochastic optimizers (e.g., SGD or Adam), as suggested in~\citep{entropysgd,PARLE}, which inherently embeds the smoothing through repeated updates.

\begin{wrapfigure}{r}{0.35\textwidth} 
\vspace{-0.35in}
\centering
\begin{minipage}{\linewidth}
  \centering
  \setlength{\tabcolsep}{3pt}
  \begin{tabular}{l c c}
    \toprule
    \textbf{Method} & \textbf{CNN-5} & \textbf{ResNet-18} \\
    \midrule
    \rowcolor{gray!20}
    \textbf{LSAM}   & \textbf{4.62} & \textbf{3.04} \\
    \textbf{LSGD}   & 5.47          & 3.75 \\
    \textbf{EASGD}  & 5.48          & 3.43 \\
    \textbf{DP-SGD} & 6.25          & 3.63 \\
    \textbf{DP-SAM} & 4.84          & 3.49 \\
    \bottomrule
  \end{tabular}
  \captionof{table}{Lowest test errors (\%) on SVHN.}
  \label{tab:final_test_errors_svhn}

  \vspace{5pt} 

  \includegraphics[width=\linewidth]{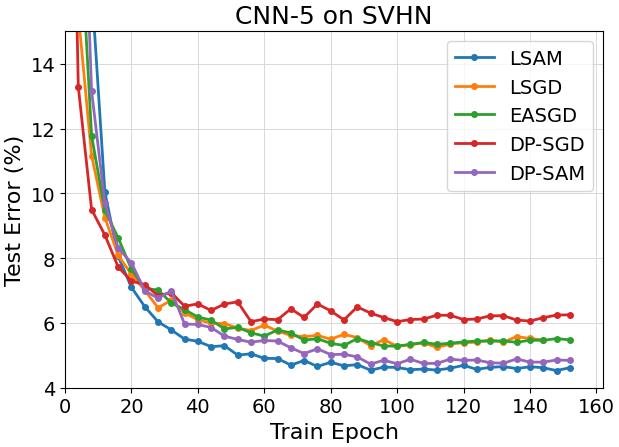}
  \includegraphics[width=\linewidth]{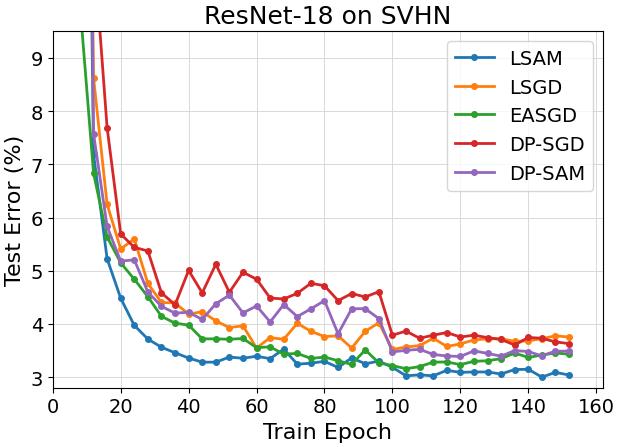}
  \captionof{figure}{Test error versus training epoch on SVHN. The zoomed plots are in \Cref{fig:svhn}, \Cref{app:exp}.}
  \label{fig:final_test_errors_svhn}
\end{minipage}
\vspace{-0.25in}
\end{wrapfigure}

\vspace{-0.05in}
\subsection{Acceleration Techniques for Optimization}
\vspace{-0.05in}
Modern optimization has developed accelerated methods that substantially improve convergence, evolving along a consistent trajectory, illustrated by shifts Momentum SGD to Nesterov's Accelerated Gradient Descent (AGD)~\citep{nesterov1983method}, Adam~\citep{KingBa15} to Adan~\citep{xie2024adan}, and ISTA~\citep{daubechies2004iterative} to FISTA~\citep{beck2009fast}. We incorporate their \emph{look-ahead} gradient characteristics via the surrogate direction:
\begin{equation}
\label{eq:gx}
g_t' =  \nabla_{y}\log \pi_{\text{LSAM}}(y_t) = -\mathbb{E}_{x\sim q(\cdot|y_t)}[\nabla_{y}k(x,y_t)]
\end{equation}
followed by the Adan-AGD-style update~\citep{xie2024adan}
$
\label{eq:agd-adan}
g_t \gets g_t' + \beta \cdot (g_t' - g_{t-1}'),
$
where $\beta$ is the momentum parameter. The momentum-augmented $g_t$ proactively anticipates future gradients through the change of gradient $(g_t' - g_{t-1}')$.

\vspace{-0.05in}
\subsection{Distributed Training}
\vspace{-0.05in}
We use Nesterov momentum~\citep{nesterov1983method} in both the inner sampling loop and the outer optimization step. The gradient $g_{t_y}$ is obtained directly from the aggregation of the asynchronously generated samples from all workers. This setup constitutes a two-time-scale stochastic approximation~\citep{ttsa,Chen_Xu_Zhang_2025}. 

Our algorithm supports fully parallel execution, enabling scalable distributed training. Each worker independently generates samples in parallel, adhering to the conditional distribution in \Cref{eq:cond_esam}. After accumulating $n \tau$ iterations across all workers, synchronization occurs to compute the LSAM score according to \Cref{eq:cond_esam} and deliver the resulting gradient to the optimizer. This forms the complete workflow of distributed LSAM, detailed in \Cref{alg:esam}.

\vspace{-0.1in}
\section{Experiments}
\vspace{-0.1in}
We compare LSAM with LSGD \citep{lsgd}, EASGD \citep{easgd}, and the standard data-parallel SGD and SAM method. For empirical experiments, we leverage the SVHN~\citep{Netzer2011ReadingDI}, CIFAR-10 and CIFAR-100 datasets \citep{krizhevsky2009learning}. On SVHN, we employ CNN-5 (a 5-layer convolutional neural network \citep{lecun1998gradient,krizhevsky2012imagenet}) and ResNet-18 \citep{he2016deep}. For CIFAR-10, assessments are performed with VGG-16 \citep{simonyan2014very}, ResNet-20 and WRN-16×10 \citep{zagoruyko2016wide}; and for CIFAR-100, we utilize ResNet-34, WRN-28×10 and WRN-40×10.

\vspace{-0.1in}
\subsection{Distributed Architecture}
\vspace{-0.1in}

We use a distributed architecture the same as \citep{teng2022leaderstochasticgradientdescent,GRAWA}, featuring a global server and several local servers that synchronize via \texttt{GLOO} for CPU coordination, alongside workers employing \texttt{NCCL} for GPU communication. This architecture only requires naive PyTorch with its build-in distributed training library.
This hierarchical configuration facilitates scalable multi-machine, multi-GPU training, with the global server managing aggregations across machines, local servers overseeing intra-node activities, and workers executing parallel GPU operations, accommodating both \emph{asynchronous} and \emph{synchronous} updates. Further details are in~\Cref{app:dist_training}.
\vspace{0.05in}

\begin{wrapfigure}{r}{0.65\textwidth} 
\vspace{-0.4in}
\centering
\begin{minipage}{0.65\textwidth}
  \begin{subfigure}[t]{0.49\linewidth}
    \includegraphics[width=\linewidth]{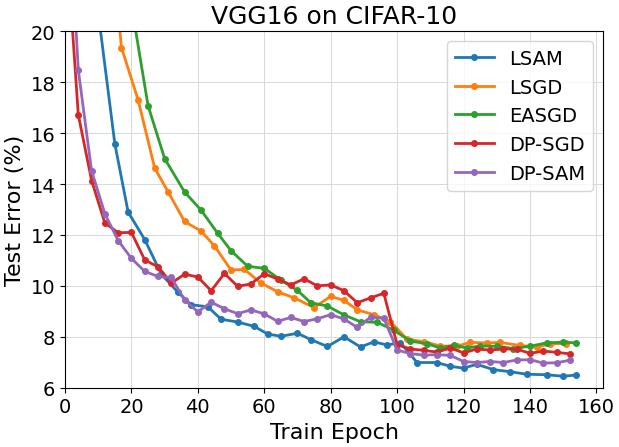}
  \end{subfigure}\hfill
  \begin{subfigure}[t]{0.49\linewidth}
    \includegraphics[width=\linewidth]{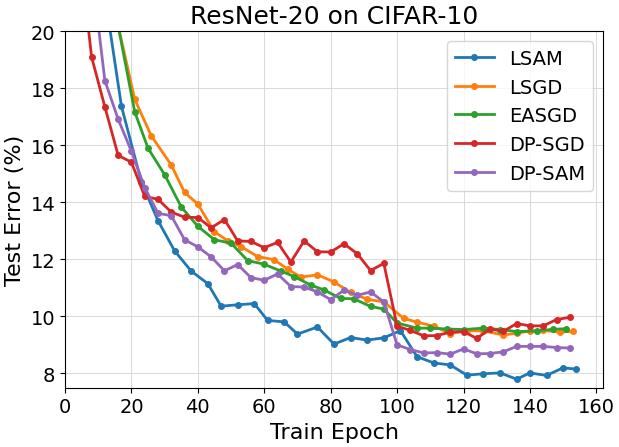}
  \end{subfigure}

  \begin{subfigure}[t]{0.49\linewidth}
    \includegraphics[width=\linewidth]{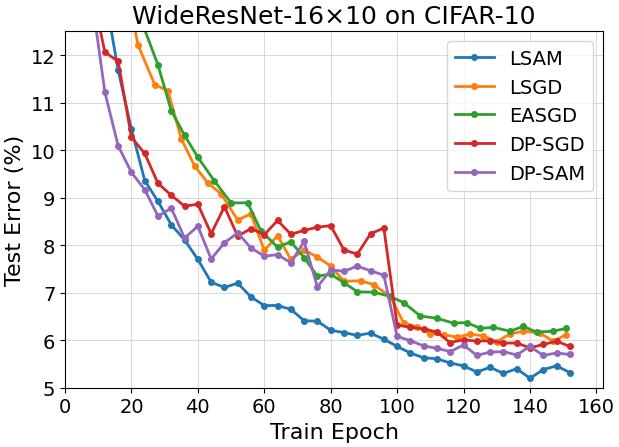}
  \end{subfigure}\hfill
  \begin{subfigure}[t]{0.49\linewidth}
    \includegraphics[width=\linewidth]{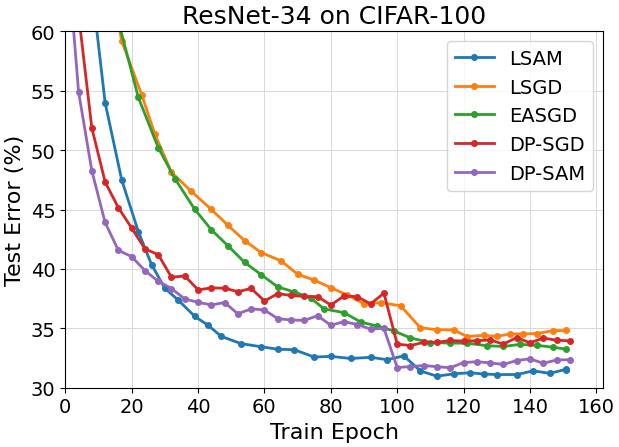}
  \end{subfigure}

  \begin{subfigure}[t]{0.49\linewidth}
    \includegraphics[width=\linewidth]{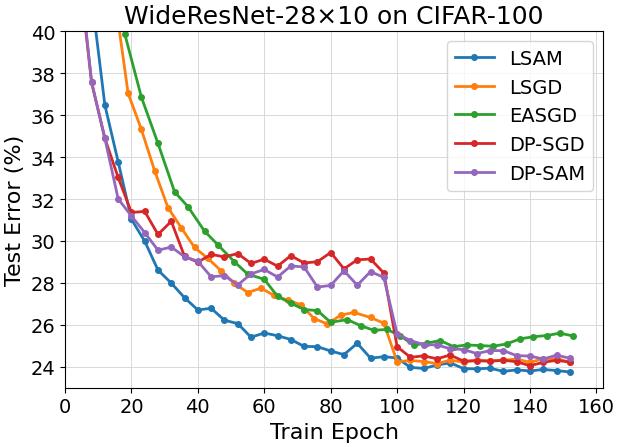}
  \end{subfigure}\hfill
  \begin{subfigure}[t]{0.49\linewidth}
    \includegraphics[width=\linewidth]{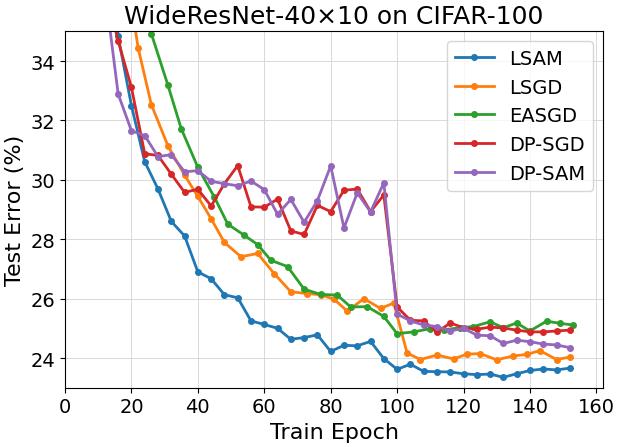}
  \end{subfigure}
  \vspace{-0.05in}
  \caption{Test error versus training epoch on CIFAR-10 and CIFAR-100 datasets. The zoomed plots are in \Cref{fig:cifar10_cifar100}, \Cref{app:exp}.}
  \label{fig:final_test_errors_cifar}
\end{minipage}
\vspace{-0.25in}
\end{wrapfigure}

\vspace{-0.1in}
\subsection{Experiment Setup}
\vspace{-0.1in}
For baselines, we utilize Pytorch's DataParallel~\citep{NEURIPS2019_9015} for SGD and SAM, denoted as DP-SGD and DP-SAM, respectively, which synchronously aggregate gradients from all workers for each iteration. We also incorporate asynchronous versions of EASGD and LSGD, which perform local updates for a few steps before synchronization.

More specifically, for DP-SGD and DP-SAM, the communication interval is $\tau = 1$. In contrast, EASGD, LSGD, and our proposed LSAM operate asynchronously with a communication interval of $\tau = 16$. Across all SVHN, CIFAR-10 and CIFAR-100 experiments, we utilize $n = 4$ workers, each with a per-worker batch size of 128. Details of hyperparameter tuning are in \Cref{app:hyper}. All methods are trained for $150$ epochs, with the learning rate reduced by a factor of 0.1 at the $100${\small th} epoch, aligning with with a reduction in test error. 
\vspace{-0.08in}

\begin{table*}[t]
\centering
\setlength{\tabcolsep}{5pt}
\begin{tabular}{l c c c c c c}
\toprule
\multicolumn{1}{c}{} & \multicolumn{3}{c}{\textbf{CIFAR-10}} & \multicolumn{3}{c}{\textbf{CIFAR-100}} \\
\cmidrule(lr){2-4} \cmidrule(lr){5-7}
\textbf{Method} & \textbf{VGG-16} & \textbf{ResNet-20} & \textbf{WRN-16$\times$10} & \textbf{ResNet-34} & \textbf{WRN-28$\times$10} & \textbf{WRN-40$\times$10} \\
\midrule
\rowcolor{gray!20} 
\textbf{LSAM}   & \textbf{6.50} & \textbf{8.15} & \textbf{5.32} & \textbf{31.46} & \textbf{23.75} & \textbf{23.66} \\
\textbf{LSGD}   & 7.72          & 9.48          & 6.12          & 34.82          & 24.22          & 24.03 \\
\textbf{EASGD}  & 7.76          & 9.57          & 6.25          & 33.24          & 25.47          & 25.11 \\
\textbf{DP-SGD} & 7.33          & 9.97          & 5.87          & 33.95          & 24.22          & 24.94 \\
\textbf{DP-SAM} & 7.08          & 8.89          & 5.70          & 32.36          & 24.41          & 24.35 \\
\bottomrule
\end{tabular}
\vspace{-0.05in}
\caption{Lowest test errors (\%) on CIFAR-10 and CIFAR-100 datasets.}
\label{tab:final_test_errors_cifar}
\vspace{-0.2in}
\end{table*}

\subsection{Results}
\vspace{-0.05in}
The lowest test errors across workers and the error–epoch curves for SVHN are shown in \Cref{tab:final_test_errors_svhn} and \Cref{fig:final_test_errors_svhn}, where LSAM clearly outperforms all baselines. This trend is reinforced in \Cref{tab:final_test_errors_cifar} and \Cref{fig:final_test_errors_cifar}, highlighting LSAM’s markedly faster convergence. Across all dataset–architecture settings, LSAM consistently achieves lower final error and faster convergence than competing methods.


\vspace{-0.1in}
\section{Conclusion}
\vspace{-0.1in}
We introduce LSAM, a reformulation of SAM that unifies its interpretation as both an optimization and a sampling procedure, specifically tailored for asynchronous distributed training. LSAM achieves faster convergence and improved generalization, while reducing communication overhead and preserving the theoretical guarantees and empirical advantages of SAM.

\clearpage
\bibliographystyle{plainnat}
\bibliography{neurips_2025, bib/sampling, bib/optimizing, bib/datasets, bib/parallel}

\clearpage
\appendix
\onecolumn

\begin{center}
\noindent\rule{\textwidth}{4pt}
  \vskip 0.02in
{\LARGE\bf LSAM: Asynchronous Distributed Training with Landscape-Smoothed Sharpness-Aware Minimization for Improved Generalization (Appendix) \par}
  \vskip -0.05in%
\noindent\rule{\textwidth}{1pt}
\end{center}

\section{Additional Theorems}
\label{app:additional_theorems}

\setcounter{assumption}{0}
\begin{assumption} \label{thm:ass_fx}
Let $f:\mathbb{R}^{d}\to\mathbb{R}$ be a continuously differentiable
potential satisfying\footnote{%
Typical sufficient conditions are
$f(x)\to\infty$ as $\lVert x\rVert\to\infty$
or $f(x)\ge c\lVert x\rVert^{q}$ for some $q>1$.}
\begin{equation}\label{eq:boltzmann-int}
  Z_0 \;:=\; \int_{\mathbb{R}^{d}} e^{-f(x)}\,\mathrm{d}x < \infty .
\end{equation}
Fix parameters $\rho>0$ and $\gamma>0$ and define the (bounded)
\emph{look-back map} as
\begin{equation}\label{eq:look_back_map}
  T_{\rho,\gamma}(x)
  \;=\;
  x+\frac{\rho\,\nabla f(x)}{\lVert\nabla f(x)\rVert+\gamma},
  \qquad x\in\mathbb{R}^{d},
\end{equation}
together with its \emph{unnormalised density}
\[
  \tilde\pi_{\rho,\gamma}(x)
  \;:=\;
  \exp\bigl[-f\bigl(T_{\rho,\gamma}(x)\bigr)\bigr],
  \;
  Z_{\rho,\gamma}
  \;:=\;
  \int_{\mathbb{R}^{d}}\tilde\pi_{\rho,\gamma}(x)\,\mathrm{d}x.
\]
Additionally, either of the following conditions holds:
\begin{enumerate}[label=\textup{(A\arabic*)}, labelindent=1em, leftmargin=2.5em, noitemsep]
  \item \(f\) is convex, or
  \item \(f\) is \textit{globally \(L\)-Lipschitz} for some \(L > 0\), i.e.,  
        \(\lvert f(y) - f(x) \rvert \leq L \norm{y - x}\) for all \(x, y \in \R^d\),
\end{enumerate}
\end{assumption}

\begin{proposition}[Existence of normalized density]
\label{thm:lookback-density}
Under Assumption \ref{thm:ass_fx}, 
one has $0<Z_{\rho,\gamma}<\infty$.
Consequently
\begin{equation}
  \pi_{\text{SAM}}(x) = \pi_{\rho,\gamma}(x)
  \;:=\;
Z_{\rho,\gamma}^{-1}\,\tilde\pi_{\rho,\gamma}(x)
\end{equation}
is a well-defined probability density on~$\mathbb{R}^{d}$. 
\end{proposition}

\begin{remark}[Sharper concentration under convexity]\label{rmk:convex_sharper}
Inequality \eqref{eq:convex_bound} implies \(Z_{\rho,\gamma} \leq Z_0\) when \(f\) is convex. Thus, the tails of \(\pi_{\rho,\gamma}\) never decays
\emph{slower} than that of the baseline Gibbs density \(e^{-f}/Z_0\), 
meaning \(\pi_{\rho,\gamma}\) is more concentrated. The parameter $\gamma$ prevents division by zero at critical points of~$f$
(where the gradient has norm zero).
\end{remark}

To construct LSAM objective function, we define the target distribution as the convolution between the sharpness-aware distribution and a kernel-based Gibbs distribution:
\vspace{-0.1in}
\begin{equation}
\label{eq:esam}
\pi_{\text{LSAM}}(y) \propto \int \pi_{\text{SAM}}(x) \, \pi_{\text{kernel}}(x,y) \, dx.
\end{equation}

\vspace{-0.12in}
Here, $\pi_{\text{kernel}}(x,y) = \exp(-k(x,y))$ denotes a kernel that encodes similarity between parameters $x$ and $y$, with $k(x,y)$ typically a positive semi-definite function (e.g., Gaussian) that decays with distance, thereby facilitating local smoothing by emphasizing nearby points in the parameter space.

\begin{assumption}[Kernel slice properties]\label{ass:kernel_slice}
Let \((\mathcal X,\mathcal B,dx)\) be a $\sigma$-finite measure space and  
let \(k:\mathcal X\times\mathcal X\to[0,\infty)\) be a measurable function such that  

\begin{enumerate}[label=\textup{(B\arabic*)}, labelindent=1em, leftmargin=2.5em, noitemsep]
\item \textbf{Non-negativity:}\; \(k(x,y)\ge 0\) for all \(x,y\in\mathcal X\).
\item \textbf{Finite, $x$-independent normalizer:}\;
      \[
        Z\;:=\;\int_{\mathcal X} \exp\left(-k(x,y)\right)\,dy \;\in\;(0,\infty) 
      \]
      $\text{for every }x \in \mathcal X$ with the value \(Z\) \emph{independent} of \(x\).
\end{enumerate}
\end{assumption}

The following proposition demonstrates that a family of stationary kernels, including gaussian kernels and $\alpha$-stable noise kernels, satisfies \Cref{ass:kernel_slice}.

\newcommand{\Folland}{\citep[Section~2.7]{folland1999real}}
\begin{corollary}[Tail condition ensuring a finite normalizer; refer to \Folland]
\label{cor:int_phi}
Let $\phi:\R^{d}\to[0,\infty)$ be measurable and define the stationary
kernel
\[
      k(x,y)\;=\;\phi(x-y),\qquad x,y\in\R^{d}.
\]
If $z:=x-y$ and
\begin{equation}
      0 \;<\; \int_{\R^{d}} e^{-\phi(z)}\,dz \;<\;\infty,
      \label{eq:int_finite}
\end{equation}
then the normalizing factor
\(
      Z(x)=\int_{\R^{d}} e^{-k(x,y)}\,dy
\)
is finite, strictly positive, and independent of \(x\). A simple pair of \emph{sufficient} tail-growth assumptions guaranteeing
\eqref{eq:int_finite} are
\begin{enumerate}[label=\emph{(\roman*)}, leftmargin=2.4em]
\item (\textbf{Polynomial/exponential growth})  
      there exist constants \(c>0,\ \alpha>0\) such that  
      \(\phi(z)\ge c\,\|z\|^{\alpha}\) for all sufficiently large \(z\);
\item (\textbf{Super-logarithmic growth})  
      there exist \(\varepsilon>0\) and \(R>0\) with  
      \(\displaystyle
            \phi(z)\;\ge\;(d+\varepsilon)\,\log\bigl(1+\|z\|\bigr)
      \;\text{for }\|z\|\ge R.\)
\end{enumerate}
Either condition implies \(e^{-\phi}\in L^{1}(\R^{d})\).
\end{corollary}

\begin{remark}
Classical examples satisfying the proposition include the exponential–power family
$\phi(z)= \lambda\lVert z\rVert^{\alpha}$
(Gaussian when $\alpha=2$), the Matérn class, the Cauchy kernel, and any integrable compact-support radial kernel. Non-stationary kernels, including linear, polynomial, or sigmoid kernels typically \emph{do not} yield a $x$-independent normalizer and thus require an alternative normalization approach.
\end{remark}

We proceed to demonstrate that $\pi_{\text{SAM}}$ is a valid probability distribution and derive the gradient of $\log \pi_{\text{LSAM}}$, referred to as the \textit{score}, as presented in \Cref{thm:kernel_density} and \Cref{lma:score}.

\begin{proposition}[Kernel-modulated density]\label{thm:kernel_density}
Assume $\mathcal{X} = \mathbb{R}^d$ and \Cref{thm:ass_fx,ass:kernel_slice} hold.  
Define, for \(y\in\mathcal X\),
\[
  \pi_{\text{LSAM}}(y) \;:=\;
  \frac{\displaystyle
        \int_{\mathcal X} 
          e^{-f ( T_{\rho,\gamma} (x) )}\,e^{-k(x,y)}\,dx}
       {Z_{\rho,\gamma} \,Z}.
\]
Then \( \pi_{\text{LSAM}} \) is a probability density on \((\mathcal X,\mathcal B,dy)\); that is,
\(\pi_{\text{LSAM}}(y)\ge 0\) for every \(y\) and \(\int_{\mathcal X} \pi_{\text{LSAM}}(y)\,dy = 1\).
\end{proposition}

We would like understand the benefits of the $\pi_{\text{LSAM}}$ formulation. For comparison, we group PARLE~\citep{PARLE}, EASGD \citep{easgd} and Entropy-SGD \citep{entropysgd} under the label \textit{ESGD}, due to their shared mathematical foundation for solving optimization problems. The distributions of SAM and ESGD are denoted as $\pi_{\text{SAM}}$ and $\pi_{\text{ESGD}}$, respectively.

\begin{wrapfigure}{r}{0.61\textwidth} 
  \centering
  \vspace{-0.15in} 
  \includegraphics[width=\linewidth]{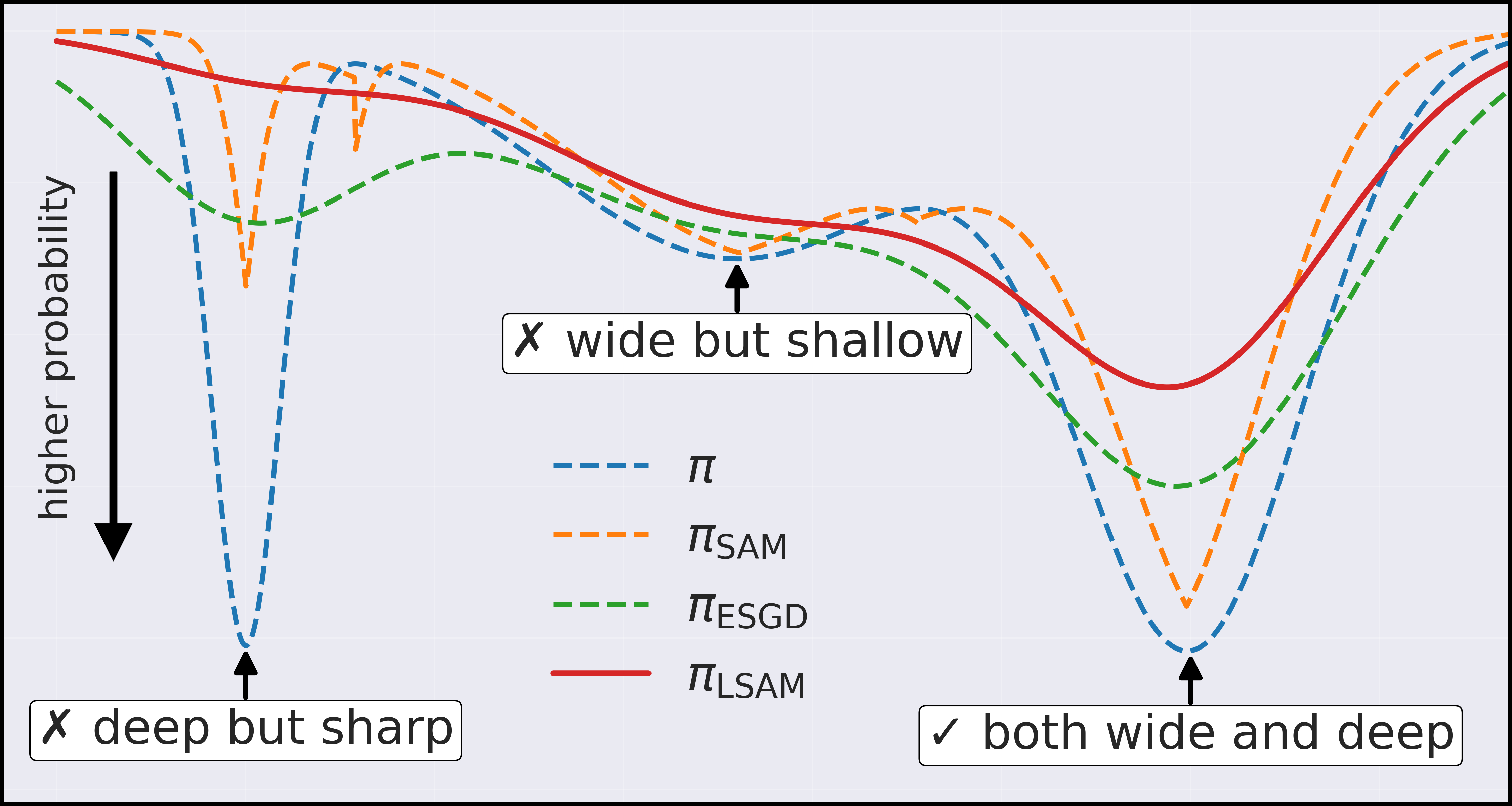}
  \vspace{-0.16in}
  \caption{Probability landscapes associated with $\pi$, $\pi_{\mathrm{SAM}}$, 
  $\pi_{\mathrm{ESGD}}$, and $\pi_{\mathrm{LSAM}}$. Among the three convergence 
  modes—deep but sharp, wide but shallow, and both wide and deep—ESGD may converge 
  to the first, SAM to the second, whereas LSAM distinctively achieves the third.}
  \label{fig:cv}
  \vspace{-0.25in}
\end{wrapfigure}

\textbf{Motivation} As shown in \Cref{fig:cv}, $\pi_{\text{SAM}}$ suppresses probability mass at sharp minima regardless of depth, $\pi_{\text{ESGD}}$ spreads mass outward to concentrate at deep minima regardless of width, while $\pi_{\text{LSAM}}$ achieves a balanced, smoother, and more selective landscape that empirically yields superior generalization. Consequently, both $\pi_{\text{SAM}}$ and $\pi_{\text{LSAM}}$ can circumvent sharp minima, whereas $\pi_{\text{ESGD}}$, although intended to seek flat minima, could still converge to deep yet sharp regions. 

These distributional differences drive parameters toward distinct basins, motivating optimization under $\pi_{\text{LSAM}}$ to target minima that are both wide and deep. Sampling from the proposed distribution is typically attainable using methods like those described in \citep{max2011,li2024entropymcmc}. These method assembles first-order gradient descent techniques. From an optimization standpoint, $\pi_{\text{LSAM}}$ creates a smoother landscape that is less likely to stall at shallow minima, while $\pi_{\text{SAM}}$ remains at risk of getting trapped in such sub-optimal basins. 

\section{Proofs}
\label{app:proofs}
\subsection{Proofs for \Cref{app:additional_theorems}}
\begin{proof}[Proof of \Cref{thm:lookback-density}]
We take the following steps to show that $0<Z_{\rho,\gamma}<\infty$ where 
$Z_{\rho,\gamma} = \int_{\mathbb{R}^{d}}\tilde\pi_{\rho,\gamma}(x)\,\mathrm{d}x$.

\noindent \textit{Step 1: Positivity and measurability.}  
Since \(\eps > 0\), the denominator \(\norm{\nabla f(x)} + \eps\) is nonvanishing. As \(f\) is \(C^1\), \(\nabla f\) is continuous, making \(T_{\rho,\eps}\) a composition of continuous maps and thus continuous (hence Borel-measurable). The exponential \(\exp(\cdot)\) is continuous and positive, so \(\tilde{\pi}_{\rho,\eps}(x) > 0\) and Borel-measurable for all \(x\).

\smallskip
\noindent\textit{Step 2: Uniform bound on the shift.}  
Define \(\delta(x) := T_{\rho,\eps}(x) - x\). Then
\begin{equation}\label{eq:delta_bound}
\norm{\delta(x)} = \rho \frac{\norm{\nabla f(x)}}{\norm{\nabla f(x)} + \gamma} \leq \rho \quad \forall x \in \R^d.
\end{equation}

\smallskip
\noindent\textit{Step 3: Integrability.} We consider the two cases separately.

\paragraph*{Case (A1): \(f\) convex.}  
By convexity, for all \(x \in \R^d\),
\[
f\left(x + \delta(x)\right) \geq f(x) + \nabla f(x)^\top \delta(x).
\]
Substituting \(\delta(x)\) and simplifying,
\[
\nabla f(x)^\top \delta(x) = \frac{\rho \norm{\nabla f(x)}^2}{\norm{\nabla f(x)} + \gamma} \geq 0,
\]
so \(f\left(x + \delta(x)\right) \geq f(x)\). Thus,
\begin{equation}\label{eq:convex_bound}
0 < \tilde{\pi}_{\rho,\gamma}(x) = \exp\left(-f\left(x + \delta(x)\right)\right) \leq e^{-f(x)}.
\end{equation}
Integrating \eqref{eq:convex_bound} and using finiteness of \(Z_0\),
\[
0 < Z_{\rho,\gamma} \leq \int_{\R^d} e^{-f(x)}  dx = Z_0 < \infty.
\]

\paragraph*{Case (A2): \(f\) globally \(L\)-Lipschitz.}  
By the Lipschitz condition and \eqref{eq:delta_bound},
\[
\left\lvert f\left(x + \delta(x)\right) - f(x) \right\rvert \leq L \norm{\delta(x)} \leq L \rho.
\]
This implies \(f\left(x + \delta(x)\right) \geq f(x) - L\rho\), so
\begin{equation}\label{eq:lipschitz_bound}
\tilde{\pi}_{\rho,\gamma}(x) \leq \exp\left(-f(x) + L\rho\right) = e^{L\rho} e^{-f(x)}.
\end{equation}
Integrating \eqref{eq:lipschitz_bound} and using finiteness of \(Z_0\),
\[
0 < Z_{\rho,\gamma} \leq e^{L\rho} Z_0 < \infty.
\]

\smallskip
\noindent\textit{Step 4: Normalization.}  
In both cases, \(0 < Z_{\rho,\gamma} < \infty\). Thus, \(\pi_{\rho,\gamma}(x)\) is non-negative, measurable, and integrates to 1.
\end{proof}

\begin{proof}[Proof of \Cref{cor:int_phi}]
We prove that each condition implies \( \int_{\mathbb{R}^d} e^{-\phi(z)}  dz < \infty \). For both cases, we split the integral:
\[
\int_{\mathbb{R}^d} e^{-\phi(z)}  dz = \int_{\|z\| < R} e^{-\phi(z)}  dz + \int_{\|z\| \geq R} e^{-\phi(z)}  dz,
\]
where \( R > 0 \) is chosen based on the condition. The first integral is always finite since \( \|z\| < R \) is compact and \( e^{-\phi(z)} \leq 1 \):
\[
\int_{\|z\| < R} e^{-\phi(z)}  dz \leq \text{Vol}(B(0,R)) < \infty.
\]
We now prove finiteness of the second integral for each condition.

\noindent \textbf{Condition (i):} Assume \(\exists c > 0, \alpha > 0, R_0 > 0\) such that \(\phi(z) \geq c \|z\|^\alpha\) for \(\|z\| \geq R_0\). Take \(R = R_0\). Then:
\[
\int_{\|z\| \geq R} e^{-\phi(z)}  dz \leq \int_{\|z\| \geq R} e^{-c\|z\|^\alpha}  dz = S_{d-1} \int_R^\infty r^{d-1} e^{-c r^\alpha}  dr.
\]
Substitute \(t = c r^\alpha\), so \(r = (t/c)^{1/\alpha}\), \(dr = \frac{1}{\alpha} c^{-1/\alpha} t^{1/\alpha-1} dt\):
\[
\int_R^\infty r^{d-1} e^{-c r^\alpha}  dr = \frac{1}{\alpha} c^{-d/\alpha} \int_{c R^\alpha}^\infty t^{d/\alpha - 1} e^{-t}  dt.
\]
The integral converges as \(t^{d/\alpha - 1} e^{-t}\) is integrable at \(\infty\) (upper incomplete gamma function).

\noindent \textbf{Condition (ii):} Assume \(\exists \varepsilon > 0, R > 0\) such that \(\phi(z) \geq (d + \varepsilon) \log(1 + \|z\|)\) for \(\|z\| \geq R\). Then:
\[
e^{-\phi(z)} \leq (1 + \|z\|)^{-(d+\varepsilon)} \quad \text{for} \quad \|z\| \geq R,
\]
so
\[
\int_{\|z\| \geq R} e^{-\phi(z)}  dz \leq S_{d-1} \int_R^\infty r^{d-1} (1 + r)^{-(d+\varepsilon)}  dr.
\]
For \(r \geq R \geq 1\), \(1 + r \leq 2r\), so:
\[
r^{d-1}(1 + r)^{-(d+\varepsilon)} \leq r^{d-1} (2r)^{-(d+\varepsilon)} = 2^{-(d+\varepsilon)} r^{-1-\varepsilon}.
\]
Thus:
\[
S_{d-1} \int_R^\infty r^{d-1} (1 + r)^{-(d+\varepsilon)}  dr \leq S_{d-1}  2^{-(d+\varepsilon)} \int_R^\infty r^{-1-\varepsilon}  dr = \frac{S_{d-1}  2^{-(d+\varepsilon)}}{\varepsilon R^\varepsilon} < \infty.
\]

For both cases, the tail integral converges. Combined with the compact part, \(\int_{\mathbb{R}^d} e^{-\phi(z)}  dz < \infty\). Strict positivity holds if \(\phi \not\equiv \infty\) on a set of positive measure.
\end{proof}

\begin{proof}[Proof of \Cref{thm:kernel_density}]
To show that $\pi_{\text{LSAM}}(y)$ is a valid density, we first check that
it is non-negative at each $y \in \R^d$ (by the non-negativity of the integrated).
Then for {normalization}, Tonelli's theorem (applied by non-negativity) and Fubini yield:
\[
\begin{aligned}
\int_{\mathcal{X}} \pi_{\text{LSAM}}(y)\,dy 
&= \frac{1}{Z_{\rho,\gamma} Z} \int_{\mathcal{X}} e^{-f(T_{\rho,\gamma} (x) )} \underbrace{\left( \int_{\mathcal{X}} e^{-k(x,y)} \,dy \right)}_{Z}\,dx \\
&= \frac{1}{Z_{\rho,\gamma}} \int_{\mathcal{X}} e^{-f(T_{\rho,\gamma} (x) ) }\,dx 
= 1,
\end{aligned}
\]
where we use $\int_{\mathcal{X}} e^{-k(x,y)}\,dy = Z$ and $\int_{\mathcal{X}} e^{-f(T_{\rho,\gamma} (x))}\,dx = Z_{\rho,\gamma}$. Thus $\pi_{\text{LSAM}}(y)$ is a valid density.
\end{proof}

\subsection{Proofs for \Cref{sec:form}}
\begin{proof}[Proof of \Cref{lma:score}]
From the definition 
$\pi_{\text{LSAM}}(y)=I(y)/(Z_{\rho,\gamma} Z)$,
it suffices to obtain the derivative of $I(y)$ 
to compute the score of $\pi_{\text{LSAM}}$.
We next apply a classical result for differentiation under integral to compute the partial derivative of $I(y)$ with respect to each $y_i$ at $y=(y_i)_{i \leq d} \in \R^d$. 

Recall that 
$I(y) := \int_{\mathbb{R}^{d}} e^{-f(T_{\rho,\gamma} (x))-k(x,y)} \, dx$.
From the conditions of \Cref{lma:score}, we have for any $(x,y) \in \R^d \times \R^d$
\[
|\nabla_{y_i} e^{-f(T_{\rho,\gamma} (x))-k(x,y)} | = |e^{-f(T_{\rho,\gamma} (x)) -k(x,y)}
 \nabla_{y_i} k(x,y) | \leq e^{-\min_x f } |\nabla_{y_i} k(x,y) |  \leq  e^{-\min_x f } g(x),
\]
because $f ( T_{\rho,\gamma} (x) ) \geq \min_x f > - \infty$ and $k(x,y) \geq 0$. 
It follows from \citet[Theorem 2.27]{folland1999real}
(extended to unbounded domains for each $y_i \in \R$) 
that $I(y)$ is differentiable on  $\R^d$, and 
\begin{align}
\nabla_y I(y) &= - \int_{\R^{d}} e^{-f(T_{\rho,\gamma} (x)) -k(x,y)}
 \nabla_{y} k(x,y) \, dx  \nonumber \\
&= -I(y)\,\E_{q}[\nabla_{y}k(x,y)]. \label{eq:gradI}
\end{align}
From \eqref{eq:gradI} and the definition of $q$, we obtain
\[
\nabla_{y}\log \pi_{\text{LSAM}}(y)
= -\,\E_{x\sim q(\cdot|y)}\bigl[\nabla_{y}k(x,y)\bigr].
\]
\end{proof}

\subsection{Proofs for \Cref{sec:opt}}
\begin{proof}[Proof of \Cref{thm:convergence_esgd}]
Define $z_t = x_t - y_t$ and $G_t = \nabla f(x_t) + \lambda z_t$. Consider the Lyapunov function $\Phi_t = f(x_t) + \frac{\lambda}{2}\|z_t\|^2$. Define $F_t(x) = f(x) + \frac{\lambda}{2}\|x - y_t\|^2$. Since $f$ is $L$-smooth, $F_t$ is $L_{\lambda}$-smooth with $L_{\lambda} = L + \lambda$, satisfying
\[
\|\nabla F_t(x) - \nabla F_t(y)\| \leq L_{\lambda}\|x - y\|
\]
for all $x,y \in \mathbb{R}^d$. Applying this to $x_t$ and $x_{t+1}$:
\[
F_t(x_{t+1}) \leq F_t(x_t) + \langle \nabla F_t(x_t), x_{t+1} - x_t \rangle + \frac{L_{\lambda}}{2}\|x_{t+1} - x_t\|^2.
\]
Substituting the update $x_{t+1} - x_t = -\eta_t(g_t + \lambda z_t)$ and $\nabla F_t(x_t) = G_t$:
\[
F_t(x_{t+1}) \leq F_t(x_t) - \eta_t \langle G_t, g_t + \lambda z_t \rangle + \frac{L_{\lambda}}{2}\eta_t^2\|g_t + \lambda z_t\|^2.
\]
Taking the conditional expectation $\mathbb{E}_t[\cdot] = \mathbb{E} [ \cdot | \mathcal{F}_{t}  ] $ where $\mathcal{F}_t$ is the $\sigma$-field of $(\xi_0, \dots, \xi_{t-1})$:

\[
\mathbb{E}_t[F_t(x_{t+1})] \leq F_t(x_t) - \eta_t \|G_t\|^2 + \frac{L_{\lambda}}{2}\eta_t^2 \mathbb{E}_t[\|g_t + \lambda z_t\|^2],
\]
where we used $\mathbb{E}_t[g_t] = 
\mathbb{E} [ \nabla f (x_t; \xi_t) | \mathcal{F}_{t} ] = \nabla f(x_t)$
since $x_t$ depends only on $(\xi_0,\dots,\xi_{t-1})$ ($x_t$ is in the sigma field $\mathcal{F}_{t}$)
and
$\xi_t$ is conditionally independent of $x_t$ given $(\xi_0,\dots,\xi_{t-1})$.
Expanding the last term:
\begin{align*}
\mathbb{E}_t[\|g_t + \lambda z_t\|^2] 
&= \mathbb{E}_t[\|(g_t - \nabla f(x_t)) + G_t\|^2] \\
&= \mathbb{E}_t[\|g_t - \nabla f(x_t)\|^2] + \|G_t\|^2 + 2\langle \mathbb{E}_t[g_t - \nabla f(x_t)], G_t \rangle \\
&\leq \sigma^2 + \|G_t\|^2,
\end{align*}
by bounded variance assumption. Thus:
\[
\mathbb{E}_t[F_t(x_{t+1})] \leq F_t(x_t) - \eta_t \|G_t\|^2 + \frac{L_{\lambda}}{2}\eta_t^2 (\sigma^2 + \|G_t\|^2).
\]
Using the $y_{t+1}$ update and $\Phi_{t+1} = f(x_{t+1}) + \frac{\lambda}{2}\|x_{t+1} - y_{t+1}\|^2$:
\begin{align*}
\Phi_{t+1} 
&= F_t(x_{t+1}) - \frac{\lambda}{2} \left(1 - (1 - \alpha)^2\right) \|x_{t+1} - y_t\|^2 \\
&\leq F_t(x_{t+1}) - \frac{\lambda}{2} \alpha \|x_{t+1} - y_t\|^2,
\end{align*}
since $1 - (1 - \alpha)^2 \geq \alpha$. Taking conditional expectation:
\[
\mathbb{E}_t[\Phi_{t+1}] \leq \mathbb{E}_t[F_t(x_{t+1})] - \frac{\lambda}{2} \alpha \mathbb{E}_t[\|x_{t+1} - y_t\|^2].
\]
Combining with previous results:
\begin{align*}
\mathbb{E}_t[\Phi_{t+1}] 
&\leq \Phi_t - \eta_t \|G_t\|^2 + \frac{L_{\lambda}}{2}\eta_t^2 (\sigma^2 + \|G_t\|^2) - \frac{\lambda}{2} \alpha \mathbb{E}_t[\|x_{t+1} - y_t\|^2] \\
&\leq \Phi_t - \frac{\eta_t}{2} \|G_t\|^2 + \frac{L_{\lambda}}{2}\eta_t^2 \sigma^2,
\end{align*}
using $\eta_t \leq 1/L_{\lambda}$. Taking total expectation and summing from $t=0$ to $T-1$:
\[
\sum_{t=0}^{T-1} \frac{\eta_t}{2} \mathbb{E}[\|G_t\|^2] \leq \Phi_0 - \mathbb{E}[\Phi_T] + \frac{L_{\lambda}\sigma^2}{2} \sum_{t=0}^{T-1} \eta_t^2 \leq \Delta + \frac{L_{\lambda}\sigma^2}{2} \sum_{t=0}^{T-1} \eta_t^2,
\]
where $\Delta = \Phi_0 - \inf_x f(x)$. With $\eta_t = \eta_0 / \sqrt{t+1}$:

\[
\forall 0 \leq t \leq T-1, \eta_t \geq \frac{\eta_0 }{ \sqrt{T} }, \quad \sum_{t=0}^{T-1} \eta_t^2 \leq \eta_0^2 (1 + \log T).
\]
Thus:
\[
\frac{1}{T} \sum_{t=0}^{T-1} \mathbb{E}[\|G_t\|^2] \leq \frac{2\Delta + L_{\lambda}\sigma^2 \eta_0^2 (1 + \log T)}{\eta_0 T} \sqrt{T} = \mathcal{O}\left(\frac{\log T}{\sqrt{T}}\right),
\]
and $\lim_{t \to \infty} \mathbb{E}[\|G_t\|^2] = 0$ follows.
\end{proof}

\begin{proof}[Proof of \Cref{thm:convergence_constant_perturbation}]
\label{pf:cpsam}
Define $z_t = x_t - y_t$, $G_t = \nabla f(x_t) + \lambda z_t$, and $\Phi_t = f(x_t) + \frac{\lambda}{2}\|z_t\|^2$. Let $L_{\lambda} = L + \lambda$. The perturbation is stochastic: $\delta_t = \rho \frac{\nabla f(x_t; \xi_t)}{\|\nabla f(x_t; \xi_t)\| + \gamma}$, so $g_t$ is evaluated at $\tilde{x}_t = x_t + \delta_t$. By $L$-smoothness and $\|\delta_t\| \leq \rho$:
\[
\|\nabla f(\tilde{x}_t) - \nabla f(x_t)\| \leq L\|\delta_t\| \leq L\rho.
\]
Start from the smoothness of $F_t(x) = f(x) + \frac{\lambda}{2}\|x - y_t\|^2$:
\[
F_t(x_{t+1}) \leq F_t(x_t) + \langle \nabla F_t(x_t), x_{t+1} - x_t \rangle + \frac{L_{\lambda}}{2}\|x_{t+1} - x_t\|^2.
\]
Substitute the update $x_{t+1} - x_t = -\eta_t(g_t + \lambda z_t)$:
\[
F_t(x_{t+1}) \leq F_t(x_t) - \eta_t \langle G_t, g_t + \lambda z_t \rangle + \frac{L_{\lambda}}{2}\eta_t^2\|g_t + \lambda z_t\|^2.
\]
Take conditional expectation:
\begin{align}
\mathbb{E}_t[F_t(x_{t+1})] 
&\leq F_t(x_t) - \eta_t \langle G_t, \mathbb{E}_t[g_t] + \lambda z_t \rangle + \frac{L_{\lambda}}{2}\eta_t^2 \mathbb{E}_t[\|g_t + \lambda z_t\|^2]. 
\label{eq:thm6cond}
\end{align}

By (C3), we have $\| \nabla f( x_t + \delta_t ; \xi_t )  - \nabla f( x_t ; \xi_t) \| \leq L \| \delta_t \| \leq L  \rho$, therefore
\[
\| \mathbb{E}_t[g_t]  -  \nabla f(x_t) \| \leq L \rho 
\]
Here, we express the inner product as:
\begin{equation}
\langle G_t, \mathbb{E}_t [g_t] + \lambda z_t \rangle = \|G_t\|^2 + \langle G_t, \mathbb{E}_t [g_t] - \nabla f(x_t) \rangle.
\label{eq:thm6G}    
\end{equation}
Next, we invoke the Cauchy-Schwarz inequality to bound the second term in \eqref{eq:thm6G}:
\begin{equation}
| \langle G_t, \mathbb{E}_t [g_t] - \nabla f(x_t) \rangle |  \leq L \rho \|G_t\|.    
\label{eq:Thm6cs}
\end{equation}

Expand the variance term:
\begin{align*}
\mathbb{E}_t[\|g_t + \lambda z_t\|^2] 
&= \mathbb{E}_t[\|(g_t - \nabla f(x_t)) + (\nabla f(x_t) + \lambda z_t)\|^2] \\
&= 2\mathbb{E}_t[\|g_t - \nabla f(x_t)\|^2] + 2\|\nabla f(x_t) + \lambda z_t\|^2 \\
&\leq 2\mathbb{E}_t[\|g_t - \nabla f(x_t)\|^2] + 2 \|G_t\|_2^2
\end{align*}
To obtain an upper bound of
$\mathbb{E}_t[\|g_t -  \nabla f(x_t)  \|^2] 
$, we compute from (C3)
\begin{align*}
    \|g_t -  \nabla f(x_t)  \|^2  &= 
    \|\nabla f(x_t +\delta_t ; \xi_t) -  \nabla f(x_t)  \|^2 \\    
    &= 
    \|\nabla f(x_t +\delta_t ; \xi_t) - \nabla f(x_t  ; \xi_t)  + \nabla f(x_t  ; \xi_t)  -  \nabla f(x_t)  \|^2 \\
    &\leq 2 \|\nabla f(x_t +\delta_t ; \xi_t) - \nabla f(x_t  ; \xi_t) \|^2 + 2 \| \nabla f(x_t  ; \xi_t)  -  \nabla f(x_t)  \|^2 \\
    &\leq 2 L^2 \rho^2 + 2  \| \nabla f(x_t  ; \xi_t)  -  \nabla f(x_t)  \|^2
\end{align*}

Combine all terms:
\begin{align}
\mathbb{E}_t[\|g_t + \lambda z_t\|^2] 
&\le 4 L^2 \rho^2 + 4 \mathbb{E}_t \| \nabla f(x_t  ; \xi_t) - \nabla 
f(x_t)  \|^2 + 2 \|G_t\|_2^2 \nonumber \\ 
&\le 4 L^2 \rho^2 + 4  \sigma^2 + 2 \|G_t\|_2^2 .
\label{eq:thm6bound2}
\end{align}

\sixin{
Now we bound \eqref{eq:thm6cond} using \eqref{eq:thm6G},\eqref{eq:Thm6cs},\eqref{eq:thm6bound2}, 
}
\begin{align*}
\mathbb{E}_t[F_t(x_{t+1})] 
&\leq F_t(x_t) - \eta_t \|G_t\|^2 + \eta_t \langle G_t, \mathbb{E}_t [g_t] - \nabla f(x_t) \rangle + \frac{L_{\lambda}}{2}\eta_t^2 \left(4 L^2 \rho^2 + 4 \sigma^2 + 2\|G_t\|_2^2\right) \\
&= F_t(x_t) - \eta_t \left(1 - L_{\lambda}\eta_t\right) \|G_t\|^2 + \eta_t L\rho \|G_t\| + 2 L_{\lambda}\eta_t^2 (L^2\rho^2 + \sigma^2).
\end{align*}
Apply Young's inequality to $\eta_t L\rho \|G_t\|$:
\[
\eta_t L\rho \|G_t\| \leq \frac{\eta_t}{4} \|G_t\|^2 + \eta_t L^2\rho^2.
\]
With $\eta_t \leq 1/(4L_{\lambda})$:
\begin{align*}
\mathbb{E}_t[F_t(x_{t+1})] 
&\leq F_t(x_t) - \eta_t \left(1 - L_{\lambda}\eta_t - \frac{1}{4}\right) \|G_t\|^2 + \eta_t L^2\rho^2 + 2 L_{\lambda}\eta_t^2 (L^2\rho^2 + \sigma^2) \\
&\leq F_t(x_t) - \frac{\eta_t}{2} \|G_t\|^2 + \eta_t L^2\rho^2 + 2 L_{\lambda}\eta_t^2 (L^2\rho^2 + \sigma^2).
\end{align*}
Relate to $\Phi_{t+1}$:
\[
\Phi_{t+1} = F_t(x_{t+1}) - \frac{\lambda}{2} \|x_{t+1} - y_t\|^2 + \frac{\lambda}{2} \|x_{t+1} - y_{t+1}\|^2 \leq F_t(x_{t+1}),
\]
Thus:
\[
\mathbb{E}_t[\Phi_{t+1}] \leq \mathbb{E}_t[F_t(x_{t+1})] \leq \Phi_t - \frac{\eta_t}{2} \|G_t\|^2 + \eta_t L^2\rho^2 + 2 L_{\lambda}\eta_t^2 (L^2\rho^2 + \sigma^2).
\]
Take total expectation and sum from $t=0$ to $T-1$:
\[
\sum_{t=0}^{T-1} \frac{\eta_t}{2} \mathbb{E}[\|G_t\|^2] \leq \Phi_0 - \mathbb{E}[\Phi_T] + L^2\rho^2 \sum_{t=0}^{T-1} \eta_t + 2 L_{\lambda}\eta_t^2 (L^2\rho^2 + \sigma^2).
\]
Multiply by 2 and set $\Delta = \Phi_0 - \inf_x f(x)$:
\[
\sum_{t=0}^{T-1} \eta_t \mathbb{E}[\|G_t\|^2] \leq 2\Delta + 2L^2\rho^2 \sum_{t=0}^{T-1} \eta_t + 4 L_{\lambda}\eta_t^2 (L^2\rho^2 + \sigma^2) \sum_{t=0}^{T-1} \eta_t^2.
\]
With $\eta_t = \eta_0 / \sqrt{t+1}$:
\[
\sum_{t=0}^{T-1} \eta_t \leq 2\eta_0 \sqrt{T}, \quad \sum_{t=0}^{T-1} \eta_t^2 \leq \eta_0^2 (1 + \log T).
\]
Thus:
\[
\sum_{t=0}^{T-1} \eta_t \mathbb{E}[\|G_t\|^2] \leq 2\Delta + 4\eta_0 L^2\rho^2 \sqrt{T} + 4 L_{\lambda} (L^2\rho^2 + \sigma^2) \eta_0^2 (1 + \log T).
\]
Since $\eta_t \geq \eta_0 / \sqrt{T}$ for $0 \leq t \leq T-1$:
\[
\sum_{t=0}^{T-1} \mathbb{E}[\|G_t\|^2] \leq \frac{\sqrt{T}}{\eta_0} \sum_{t=0}^{T-1} \eta_t \mathbb{E}[\|G_t\|^2] \leq \frac{\sqrt{T}}{\eta_0} \left[2\Delta + 4\eta_0 L^2\rho^2 \sqrt{T} + 4 L_{\lambda} (L^2\rho^2 + \sigma^2) \eta_0^2 (1 + \log T)\right].
\]
Divide by $T$:
\[
\frac{1}{T} \sum_{t=0}^{T-1} \mathbb{E}[\|G_t\|^2] \leq \frac{2\Delta}{\eta_0 \sqrt{T}} + 4 L^2\rho^2 + \frac{4 L_{\lambda} (L^2\rho^2 + \sigma^2) \eta_0 (1 + \log T)}{\sqrt{T}}.
\]
The right-hand side is $4L^2\rho^2 + \mathcal{O}\left(\frac{\log T}{\sqrt{T}}\right)$, which completes the proof.
\end{proof}

\begin{proof}[Proof of \Cref{thm:convergence_decaying_perturbation}]
Similar to the constant $\rho$ case, take total expectation and sum from $t=0$ to $T-1$:
\[
\sum_{t=0}^{T-1} \frac{\eta_t}{2} \mathbb{E}[\|G_t\|^2] \leq \Phi_0 - \mathbb{E}[\Phi_T] + L^2 \sum_{t=0}^{T-1} \eta_t \rho_t^2 + 2 L_{\lambda} \left(L^2 \sum_{t=0}^{T-1} \eta_t^2 \rho_t^2 + \sigma^2 \sum_{t=0}^{T-1} \eta_t^2 \right).
\]
Multiply by 2 and set $\Delta = \Phi_0 - \inf_x f(x)$:

\begin{align*}
\sum_{t=0}^{T-1} \eta_t \mathbb{E}[\|G_t\|^2] &\leq 2\Delta + 2L^2 \sum_{t=0}^{T-1} \eta_t \rho_t^2 + 4L_{\lambda} L^2 \sum_{t=0}^{T-1} \eta_t^2 \rho_t^2 + 4L_{\lambda} \sigma^2 \sum_{t=0}^{T-1} \eta_t^2 \\
&= 2\Delta + 2L^2 \eta_0 \rho_0^2 \sum_{t=0}^{T-1} (t+1)^{-3/2} + 4L_{\lambda} L^2 \eta_0^2 \rho_0^2 \sum_{t=0}^{T-1} (t+1)^{-2} \\
&\quad + 4L_{\lambda} \sigma^2 \eta_0^2 \sum_{t=0}^{T-1} (t+1)^{-1} \\
&\leq 2\Delta + 2L^2 \eta_0 \rho_0^2 \zeta(3/2) + 4L_{\lambda} L^2 \eta_0^2 \rho_0^2 \zeta(2) + 4L_{\lambda} \sigma^2 \eta_0^2 (1 + \log T)
\end{align*}
where $\zeta(\cdot)$ is the Riemann zeta function. Since $\eta_t \geq \eta_0 / \sqrt{T}$ for $0 \leq t \leq T-1$:
\begin{align*}
\frac{1}{T} \sum_{t=0}^{T-1} \mathbb{E}[\|G_t\|^2] 
&\leq \frac{1}{\eta_0 \sqrt{T}} \Big[ 2\Delta + 2L^2 \eta_0 \rho_0^2 \zeta(3/2) \\
&\quad + 4L_{\lambda} L^2 \eta_0^2 \rho_0^2 \zeta(2) + 4L_{\lambda} \sigma^2 \eta_0^2 (1 + \log T) \Big] \\
&= \frac{2\Delta}{\eta_0 \sqrt{T}} + \frac{2L^2 \rho_0^2 \zeta(3/2)}{\sqrt{T}} + \frac{4L_{\lambda} L^2 \eta_0 \rho_0^2 \zeta(2)}{\sqrt{T}} + \frac{4L_{\lambda} \sigma^2 \eta_0 (1 + \log T)}{\sqrt{T}}
\end{align*}
The averaged norm $\frac{1}{T} \sum_{t=0}^{T-1} \mathbb{E}[\|G_t\|^2] $ converges to zero as $T \to \infty$ with rate $\mathcal{O}\left(\frac{\log T}{\sqrt{T}}\right)$.
\end{proof}

\begin{proof}[Proof of \Cref{lem:anchor-gap}]
\emph{Step 1 (uniform bound on \(\mathbb{E}\|z_t\|\)).}
The update rule gives
\[
z_{t+1}=(1-\alpha)\bigl[(1-\lambda\eta_t)z_t-\eta_t g_t\bigr].
\]
Taking expectations and using the triangle inequality (\sixin{for LSAM with decaying
perturbation in \Cref{thm:convergence_decaying_perturbation}, $G=C + \rho_0 L$ since $\E_t [\| g_t \|] \leq \E_t [ \| \nabla f ( x_t ; \xi_t) \|] + \rho_t L \leq C + \rho_0 L $; for ESGD, $G=C$}),
\[
\mathbb{E}\|z_{t+1}\|
  \le(1-\alpha)\bigl[(1-\lambda\eta_t)\mathbb{E}\|z_t\|+\eta_t G\bigr].
\]
Inductively suppose \(\mathbb{E}\|z_t\|<D\).  Because \(G\le\lambda D\),
\[
(1-\lambda\eta_t)D+\eta_t G\le(1-\lambda\eta_t)D+\eta_t\lambda D=D,
\]
so \(\mathbb{E}\|z_{t+1}\|<(1-\alpha)D<D\).  The claim follows.

\medskip
\noindent
\emph{Step 2 (rate for \(\mathbb{E}\|z_t\|^{2}\)).}
Let \(d_t=\mathbb{E}\|z_t\|^{2}\).  Squaring the update and expanding,
\[
\|z_{t+1}\|^{2}
=(1-\alpha)^{2}\!\left[(1-\lambda\eta_t)^{2}\|z_t\|^{2}
      -2\eta_t(1-\lambda\eta_t)z_t^{\top}g_t
      +\eta_t^{2}\|g_t\|^{2}\right].
\]
Taking conditional expectations and then full expectations, using
\(\lvert z_t^{\top}\mathbb{E}_t[g_t]\rvert\le\|z_t\|\cdot G\) and
\(\mathbb{E}\|z_t\|<D\),
\[
d_{t+1}
\le(1-\alpha)^{2}(1-\lambda\eta_t)^{2}d_t
    +2(1-\alpha)^{2}\eta_t(1-\lambda\eta_t)DG
    +(1-\alpha)^{2}\eta_t^{2}G^{2}.
\]
Set 
\[
A_t=(1-\alpha)^{2}(1-\lambda\eta_t)^{2},
\qquad
F_t=(1-\alpha)^{2}\!\left[2\eta_t(1-\lambda\eta_t)DG+\eta_t^{2}G^{2}\right].
\]
Since \(\eta_t^{2}\le\eta_0\eta_t\), there exists a constant
\(K>0\) with \(F_t\le K\eta_t\).
Iterating the inequality,
\[
d_t\le\Bigl(\prod_{k=0}^{t-1}A_k\Bigr)d_0
      +\sum_{k=0}^{t-1}F_k\prod_{j=k+1}^{t-1}A_j.
\]
Write \(\rho=(1-\alpha)^{2}<1\).  Because \(A_k\le\rho\),
\[
\prod_{k=0}^{t-1}A_k\le\rho^{t},
\qquad
F_k\le K\eta_0(k+1)^{-1/2}.
\]
Splitting the geometric sum at \(k=\lfloor t/2\rfloor\) shows
\(\sum_{k=0}^{t-1}F_k\prod_{j=k+1}^{t-1}A_j=\mathcal{O}(t^{-1/2})\),
while \(\rho^{t}d_0\) decays exponentially.  Hence
\(d_t=\mathcal{O}(t^{-1/2})\) and \(d_t\to0\).
\end{proof}

\begin{proof}[Proof of \Cref{cor:grad_norm_convergence}]
Given the definitions $z_t \coloneqq x_t - y_t$ and $G_t = \nabla f(x_t) + \lambda z_t$ with asymptotic bounds $\mathbb{E} \|G_t\|^2 = \mathcal{O}\big(\frac{\log T}{\sqrt{T}}\big)$ and $\mathbb{E} \|z_t\|^2 = \mathcal{O}\big(\frac{\log T}{\sqrt{T}}\big)$, we derive the convergence rate for the gradient norm. Starting from the identity $\nabla f(x_t) = G_t - \lambda z_t$, we apply the fundamental inequality $\|a + b\|^2 \leq 2\|a\|^2 + 2\|b\|^2$ to obtain:
\[
\|\nabla f(x_t)\|^2 \leq 2\|G_t\|^2 + 2\lambda^2\|z_t\|^2.
\]
Taking expectations and substituting the given rates yields:
\[
\mathbb{E} \|\nabla f(x_t)\|^2 \leq 2\mathbb{E} \|G_t\|^2 + 2\lambda^2 \mathbb{E} \|z_t\|^2.
\]
By the asymptotic assumptions, there exist positive constants $C_1$, $C_2$ such that for sufficiently large $T$,
\[
\mathbb{E} \|G_t\|^2 \leq C_1 \frac{\log T}{\sqrt{T}} \quad \text{and} \quad \mathbb{E} \|z_t\|^2 \leq C_2 \frac{\log T}{\sqrt{T}}.
\]
Combining these bounds, we conclude:
\[
\mathbb{E} \|\nabla f(x_t)\|^2 \leq (2C_1 + 2\lambda^2 C_2) \frac{\log T}{\sqrt{T}} = \mathcal{O}\bigg(\frac{\log T}{\sqrt{T}}\bigg).
\]
\end{proof}

\section{Distributed Training}
\label{app:dist_training}
\begin{figure*}[t]
    \centering
    \includegraphics[width=\linewidth]{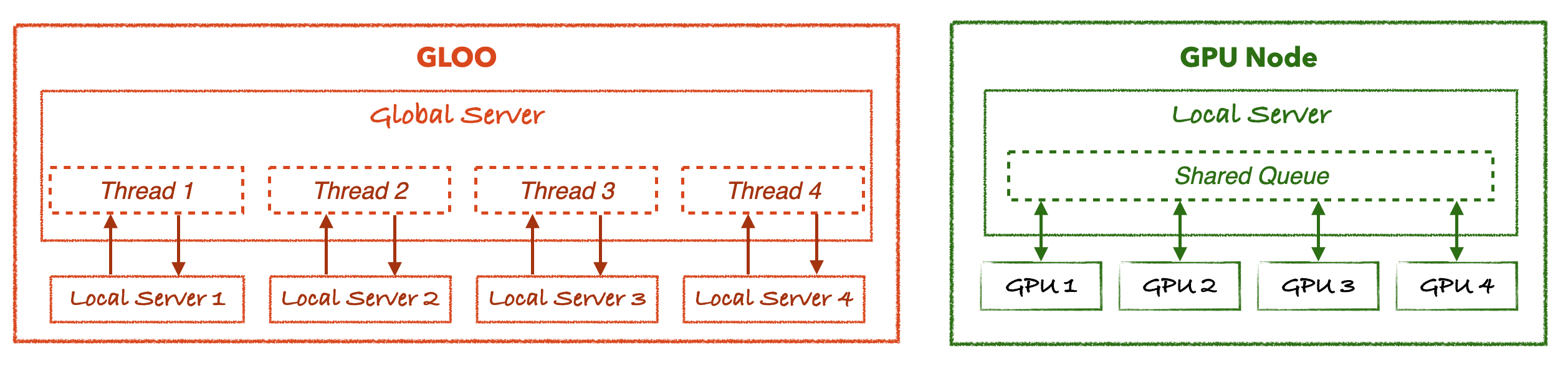}
    \caption{The global server establishes connections with each local server, while local servers communicate with their respective GPUs via thread-safe shared queues to ensure stable operations and reliable communication across the distributed system.}
    \label{fig:dist2}
\end{figure*}

\subsection{Workflow Design}
For each local server, the global server spawns a dedicated communication thread. These threads collectively maintain an iteration counter tracking total worker iterations since the last synchronization\footnote{Crucially, this counter is  implemented as a \emph{shared queue} to avoid race conditions arising from simultaneous updates by multiple threads. While Python's GIL might suggest concurrency safety, a queue proves more robust in practice. The queue structure ensures: (1) exact correspondence between worker requests and server-propagated requests, and (2) a blocking mechanism that suspends workers until their response is ready from the local server. This design avoids potential update discrepancies that could occur with shared tensor implementations.}.
At the start of every training iteration, workers send requests to their local servers and await responses. Communication occurs through the following: local servers forward requests to the global server, which evaluates system state and returns responses. The local server then enqueues these responses into worker-accessible queues, enabling workers to dequeue and execute embedded instructions—either local or distributed training—based on the global server's directives. This entire sequence, depicted in \Cref{fig:dist2}, establishes a stable workflow that orchestrates all critical component interactions.

\subsection{Batch Normalization}
A key difficulty in distributed training is managing the running mean and variance in batch normalization layers. Common strategies either let each worker maintain its own statistics (ignoring synchronization) or introduce extra coordination to sync them. There is no universally optimal approach in asynchronous settings. In our framework—which is built around loosely coupled parameters—we do not synchronize these statistics, and empirical results show no degradation in performance.

\subsection{PyTorch Data Loader}
In PyTorch, the \texttt{dataLoader} by default uses CPU multiprocessing, which can incur overhead from repeatedly spawning and tearing down worker processes. To alleviate this, we enable the \texttt{persistent workers} option so that workers remain alive after each epoch’s dataset traversal, thereby avoiding repeated initialization and improving multi-epoch efficiency.

\section{Hyperparameters}
\label{app:hyper}
\subsection{Hyperparameter Search}
We performed a grid search across two key hyperparameters for all methods: initial learning rates $\eta \in \{0.01, 0.02, 0.05, 0.1, 0.2, 0.3\}$ and adversarial factors $\rho \in \{0.1, 0.05, 0.01\}$. For LSAM, we explored the pulling strength formulation $\lambda = \frac{\lambda_0}{\eta \tau}$ with $\lambda_0 \in \{0.1, 0.2, 0.5, 0.9\}$, while fixing $\eta' = 1.0$ and momentum coefficient $\beta = 0.9$.

We note that in \Cref{alg:esam}, the learning rate $\eta$ for the LSAM sampler adheres to standard SGD practices. Conversely, the optimization phase of LSAM uses a constant learning rate of 1.0 and mainly relies on the Nesterov-momentum-based gradient prediction, parameterized by $\beta$, to enable accelerated optimization.

\subsection{Hyperparameter Selection}
\noindent To achieve optimal performance across all experiments, we consistently applied two configurations:
\begin{itemize}
    \item $\rho = 0.1$ for both LSAM and DP-SAM methods
    \item Architecture-specific learning rates: $0.02$ for CNNs and $0.2$ for ResNet, VGG, and WRN architectures across \textit{all} methods
\end{itemize}
For LSGD~\citep{lsgd} and EASGD~\citep{easgd}, in consistency with their original implementations, we maintain their default parameters.

\section{Additional Results}
\label{app:exp}
\begin{figure*}[ht]
  \centering
  \begin{subfigure}[t]{0.49\linewidth}
    \centering
    \includegraphics[width=\linewidth]{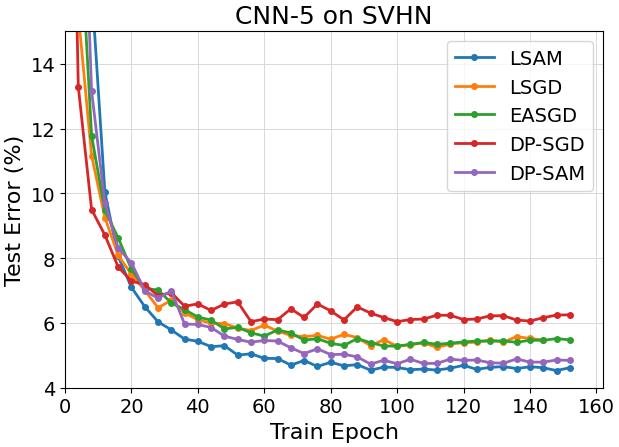}
  \end{subfigure} \hfill 
  \begin{subfigure}[t]{0.49\linewidth}
    \centering
    \includegraphics[width=\linewidth]{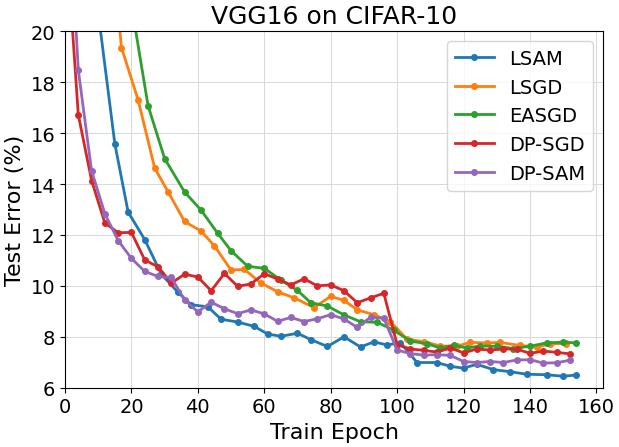}
  \end{subfigure}
  \caption{Test error versus train epoch on SVHN dataset.}
  \label{fig:svhn}
\end{figure*}

\begin{figure*}[ht]
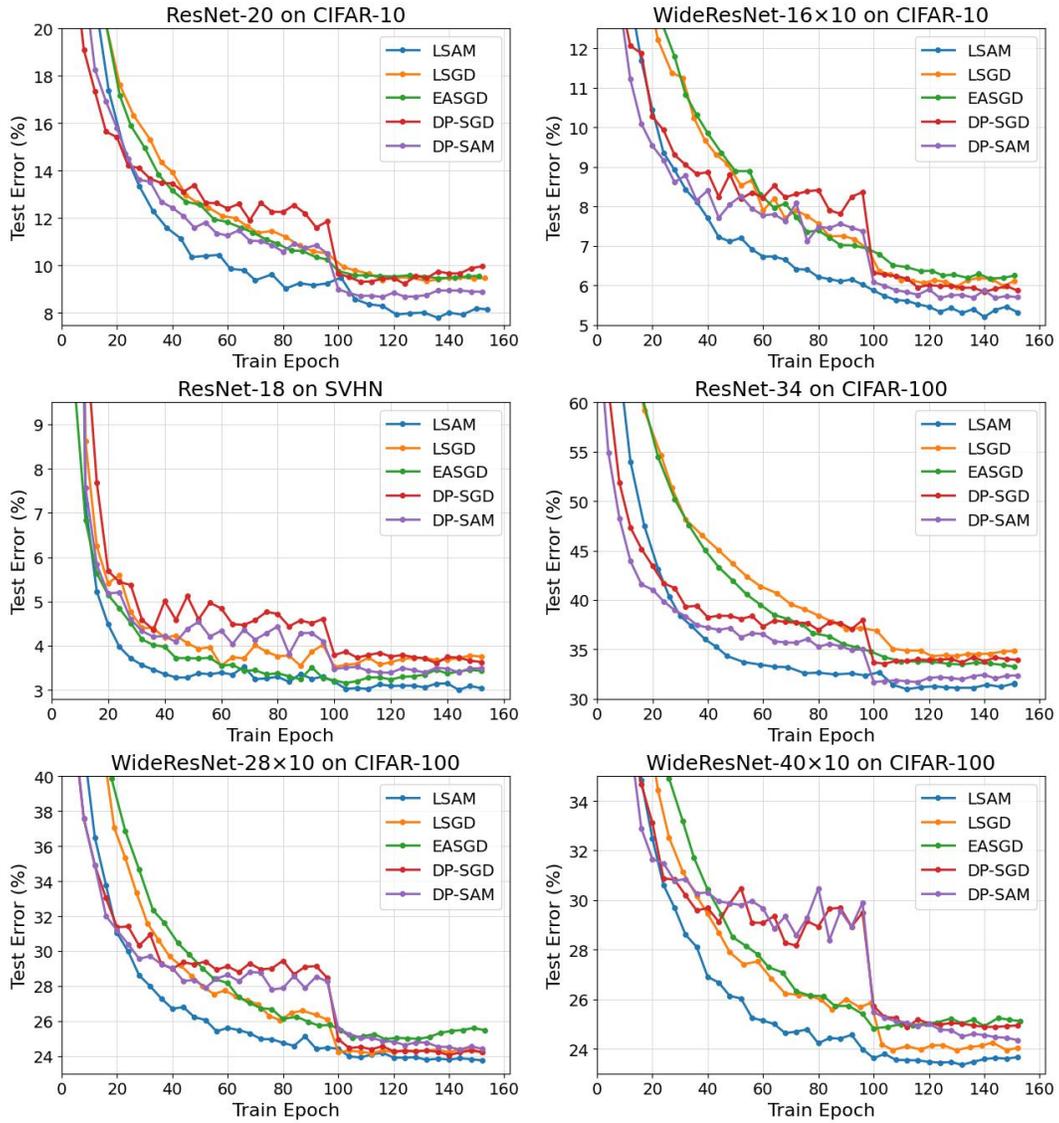

  \centering
  \begin{subfigure}[t]{0.49\linewidth}
    \centering
    \includegraphics[width=\linewidth]{figures/one_figures_lsam/test_error_cifar_resnet20_b=128_.jpg}
  \end{subfigure} \hfill 
  \begin{subfigure}[t]{0.49\linewidth}
    \centering
    \includegraphics[width=\linewidth]{figures/one_figures_lsam/test_error_cifar_wrn16x10_b=128_.jpg}
  \end{subfigure} 
  \begin{subfigure}[t]{0.49\linewidth}
    \centering
    \includegraphics[width=\linewidth]{figures/one_figures_lsam/test_error_SVHN_resnet18_b=128_.jpg}
  \end{subfigure} \hfill 
  \begin{subfigure}[t]{0.49\linewidth}
    \centering
    \includegraphics[width=\linewidth]{figures/one_figures_lsam/test_error_cifar100_resnet34_b=128_.jpg}
  \end{subfigure}
  \begin{subfigure}[t]{0.49\linewidth}
    \centering
    \includegraphics[width=\linewidth]{figures/one_figures_lsam/test_error_cifar100_wrn28x10_b=128_.jpg}
  \end{subfigure} \hfill 
  \begin{subfigure}[t]{0.49\linewidth}
    \centering
    \includegraphics[width=\linewidth]{figures/one_figures_lsam/test_error_cifar100_wrn40x10_b=128_.jpg}
  \end{subfigure}
  \caption{Test error versus train epoch on CIFAR-10 and CIFAR-100 datasets.}
  \label{fig:cifar10_cifar100}
\end{figure*}
\end{document}